\documentclass[acmsmall]{acmart}
\usepackage{booktabs,multirow,array}
\AtBeginDocument{%
  \providecommand\BibTeX{{%
    \normalfont B\kern-0.5em{\scshape i\kern-0.25em b}\kern-0.8em\TeX}}}

\setcopyright{acmcopyright}
\acmJournal{TOIS}
\acmYear{2021} \acmVolume{1} \acmNumber{1} \acmArticle{1} \acmMonth{1} \acmPrice{15.00}\acmDOI{10.1145/3457533}

\begin{document}

\title{Contextualized Knowledge-aware Attentive Neural Network: Enhancing Answer Selection with Knowledge}\thanks{The work described in this article is substantially supported supported by a grant from the Research Grant Council of the Hong Kong Special Administrative Region, China (Project Code: 14204418) and the Shenzhen General Research Project (No. JCYJ20190808182805919). Yang Deng and Yuexiang Xie contributed equally to this work. Ying Shen is the corresponding author.}

\author{Yang Deng}
\email{ydeng@se.cuhk.edu.hk}
\affiliation{%
  \institution{The Chinese University of Hong Kong}
  \city{Hong Kong}
  \postcode{999077}
}
\author{Yuexiang Xie}
\email{yuexiang.xyx@alibaba-inc.com}
\affiliation{%
  \institution{Alibaba Group}
  \country{China}
}

\author{Yaliang Li}
\affiliation{%
  \institution{Alibaba Group}
  \city{Bellevue, WA}
  \country{USA}}
\email{yaliang.li@alibaba-inc.com}

\author{Min Yang}
\affiliation{%
  \institution{SIAT, Chinese Academy of Sciences}
  \city{Shenzhen}
  \country{China}
}
\email{min.yang@siat.ac.cn}

\author{Wai Lam}
\affiliation{%
 \institution{The Chinese University of Hong Kong}
 \city{Hong Kong}}
 \email{wlam@se.cuhk.edu.hk}

\author{Ying Shen}
\affiliation{%
  \institution{Sun Yat-Sen University}
  \city{Guangzhou}
  \country{China}}
\email{sheny76@mail.sysu.edu.cn}

\renewcommand{\shortauthors}{Deng and Xie, et al.}

\begin{abstract}
Answer selection, which is involved in many natural language processing applications such as dialog systems and question answering (QA), is an important yet challenging task in practice, since conventional methods typically suffer from the issues of ignoring diverse real-world background knowledge.
In this paper, we extensively investigate approaches to enhancing the answer selection model with external knowledge from knowledge graph (KG).
First, we present a context-knowledge interaction learning framework, Knowledge-aware Neural Network (KNN), which learns the QA sentence representations by considering a tight interaction with the external knowledge from KG and the textual information.
Then, we develop two kinds of knowledge-aware attention mechanism to summarize both the context-based and knowledge-based interactions between questions and answers.
To handle the diversity and complexity of KG information, we further propose a Contextualized Knowledge-aware Attentive Neural Network (CKANN), which improves the knowledge representation learning with structure information via a customized Graph Convolutional Network (GCN) and comprehensively learns context-based and knowledge-based sentence representation via the multi-view knowledge-aware attention mechanism. 
We evaluate our method on four widely-used benchmark QA datasets, including WikiQA, TREC QA, InsuranceQA and Yahoo QA. Results verify the benefits of incorporating external knowledge from KG, and show the robust superiority and extensive applicability of our method.

\end{abstract}


\maketitle

\section{Introduction}\label{section1}
We consider the problem of answer selection, which is an active research field in natural language processing (NLP), with applications in many areas such as factoid question answering~\cite{DBLP:conf/emnlp/WangSM07,DBLP:conf/emnlp/YangYM15}, community-based question answering (CQA)~\cite{DBLP:conf/sigir/TayPLH17,DBLP:conf/aaai/DengLXCL0S20,DBLP:conf/sigir/DengZL0LS20}, and domain-specific question answering~\cite{DBLP:conf/asru/FengXGWZ15,DBLP:conf/sigir/ZhangDL20}.
Given a question and a set of candidate answers, the aim of answer selection is to determine which of the candidates answers the question accurately.

These years, various frameworks have been proposed for answer selection, ranging from traditional information-retrieval techniques for identifying the most relevant answers, which primarily focused on feature engineering, syntactic or lexical approach~\cite{DBLP:conf/emnlp/WangSM07,DBLP:conf/emnlp/SeverynM13,DBLP:conf/acl/YihCMP13}, to deep learning model based methods such as convolutional neural network (CNN)~\cite{DBLP:conf/emnlp/YangYM15,DBLP:conf/sigir/SeverynM15,DBLP:conf/emnlp/NicosiaM18} or recurrent neural network (RNN)~\cite{DBLP:conf/acl/WangN15,DBLP:conf/aaai/MuellerT16}.
However, most recent deep neural networks \cite{DBLP:conf/acl/TanSXZ16,DBLP:journals/corr/SantosTXZ16,DBLP:journals/tacl/YinSXZ16,DBLP:conf/acl/WuWS18} that achieve the state-of-the-art on the answer selection task only consider the context information of QA sentences rather than real-world background information and knowledge beyond the context, which play crucial roles in human text comprehension. Table~\ref{qaexample} lists an example question and its positive and negative answers.
In the absence of the real-world background knowledge, the negative answer may be scored higher than its positive counterpart, since the negative answer is more similar to the given question at the word level. Conversely, with background knowledge, we can correctly identify the positive answer through the relevant facts contained in a knowledge graph (KG) such as (\textit{What 's My Name}, \textit{produced\_by}, \textit{duo StarGate}), (\textit{What 's My Name}, \textit{song\_by}, \textit{Rihanna}) and even (\textit{What 's My Name}, \textit{written\_by}, \textit{Ester Dean, Traci Hale, Drake}).  

\begin{table*}
\centering
  \caption{Example of QA candidate pairs.}
  \begin{tabular}{cp{10cm}}
    \toprule
	Question & who wrote \textbf{\textit{What 's My Name}} for  \textbf{\textit{Rihanna}} ?\\
    \midrule
     Positive Answer & The R \& B song was produced by the Norwegian production \textbf{\textit{duo StarGate}} , and was written by the \textbf{\textit{duo StarGate}} along with \textbf{\textit{Ester Dean}} , \textbf{\textit{Traci Hale}} , and \textbf{\textit{Drake}} .\\
   \midrule
   Negative Answer & `` \textbf{\textit{What 's My Name}} ? '' gave \textbf{\textit{Rihanna}} her third number-one single in 2010 , as well as her eighth overall on the chart .  \\
  \bottomrule
\label{qaexample}
\end{tabular}
\end{table*}

Under this circumstance, there are two challenges remaining to be tackled in incorporating external knowledge from KG for ranking question answer pairs:

(1) The major challenge is how to introduce the large-scale factual knowledge contained in the KG into the context-based answer selection model, as well as facilitate the interactions between the context and knowledge information. Even though recent advances in constructing large-scale knowledge graphs have enabled QA systems to return an answer from a KG~\cite{DBLP:conf/acl/DongWZX15,DBLP:conf/acl/DaiLX16}, the exploration of background knowledge from KGs into QA sentence representations for natural answer selection is still a relatively new territory and under-explored ~\cite{DBLP:conf/cikm/RaoHL16,DBLP:conf/kdd/WangJY17,DBLP:conf/aaai/TayTH18}. On the other hand, previous studies~\cite{DBLP:conf/acl/YangM17,DBLP:conf/aaai/XinL0S18,DBLP:conf/aaai/Han0S18} on other NLP tasks (e.g., relation extraction and machine reading) exclusively adopt external knowledge from KG to conduct knowledge-aware learning of individual sentences. However, these works ignore the interrelations between different sentences and the semantic relationship of words and knowledge between QA pairs, which can endow the redundancy and noise with less importance and put emphasis on informative representations.

(2) Another upcoming challenge for incorporating external knowledge into answer selection is that simply adopting pre-trained knowledge embeddings falls short of modeling specific information in certain question/answer sentences. Based on the existing knowledge embedding
methods (e.g., TransE~\cite{DBLP:conf/nips/BordesUGWY13} and transH~\cite{DBLP:conf/aaai/WangZFC14}), some recent works~\cite{DBLP:conf/sigir/ShenDYLD0L18,DBLP:conf/coling/DengSYLDFL18} attempt to learn the knowledge-aware sentence representations for ranking question-answer pairs. Despite the effectiveness of these pre-trained knowledge embedding methods, the representation of knowledge is fixed and unable to be adapted to different contexts (i.e., different question/answer sentences). Therefore, we consider that the knowledge representation is supposed to be contextualized and adaptive to represent the entire question/answer sentences, and the use of semantically-fixed knowledge embeddings limits the performance of models.

To tackle these issues in answer selection, we present an extensive study on the incorporation of external knowledge from knowledge graphs into answer selection models accordingly. In specific, we propose a general framework, Knowledge-aware Neural Network (KNN), and a comprehensive model, Contextualized Knowledge-aware Attentive Neural Network (CKANN), for knowledge-aware answer selection:

(1) We develop a Knowledge-aware Neural Network (KNN) framework to introduce the external knowledge from KG for sentence representation learning.
KNN encodes the QA sentences into context-based representations, and embeds the knowledge of questions and answers in knowledge representations. 
A context-guided attentive CNN is designed to learn the knowledge-based sentence representation embedded in the discrete candidate entities in KG.
Then two kinds of knowledge-aware attention mechanisms are designed to adaptively attend the important information of questions and answers by simultaneously considering the context information and the association of knowledge beyond the text.

(2) We further propose a Contextualized Knowledge-aware Attentive Neural Network (CKANN), which learns the contextualized knowledge representation to better handle the diversity and complexity of KG information.
Within the framework of KNN, the derived knowledge sequences of sentences are mapped into an external KG to generate an Entity Graph. The contextualized knowledge representation is learned from the Entity Graph with Graph Convolutional Network (GCN)~\cite{DBLP:conf/nips/DefferrardBV16}.
Besides, we improve the knowledge-aware attention mechanism via a multi-view attention mechanism, comprehensively capturing the semantic relationships of words and knowledge between QA pairs.

According to the aforementioned discussion, our empirical analysis aims at answering the following research questions:
\begin{itemize}
    \item \textbf{RQ1:} Can answer selection benefit from incorporating external knowledge?
    \item \textbf{RQ2:} How can the context and knowledge information effectively interact with each other? 
    \item \textbf{RQ3:} Can GCN improve the contextualized knowledge representation learning?
    \item \textbf{RQ4:} Can the proposed framework be adapted to different kinds of exsiting context-based answer selection models?
    \item \textbf{RQ5:} How does the knowledge representation learning affect the overall performance?
    \item \textbf{RQ6:} How does the completeness of the knowledge graph affect he overall performance?
\end{itemize}

The main contributions of this paper can be summarized as follows: 
\begin{itemize}
    \item We present a general context-knowledge interaction learning framework, Knowledge-aware Neural Network (KNN), with two kinds of knowledge-aware attention mechanisms, to exploit external knowledge from knowledge graph and capture the background information between QA pairs for answer selection;
    \item Based on KNN framework and knowledge-aware attentive mechanism, we present an advanced model, namely Contextualized Knowledge-aware Attentive Neural Network (CKANN), which dynamically leverages the external knowledge with structure information from KGs to capture the hidden semantic relationship between QA pairs for answer selection, and comprehensively learns the interactions from different views of information via the multi-view knowledge-aware attention mechanism;
    \item The experimental results show that the proposed methods consistently outperform the state-of-the-art methods. Extensive experiments are conducted to demonstrate the effectiveness and applicability of the proposed method from various perspectives.
\end{itemize}

This article substantially extends our previous work published as a conference paper~\cite{DBLP:conf/sigir/ShenDYLD0L18}.  
We substantially improve the method with a novel contextualized knowledge-based learning module and a multi-view attention mechanism. We formalize the method as a general framework. In addition, we conduct extensive experiments to validate the proposed method from various aspects, such as the applicability of the proposed method, the effectiveness of knowledge embedding methods, and the contribution of additional features.

The rest of the article is organized as follows. Section~\ref{section2} reviews related works.
Section~\ref{section3} briefly introduces the problem definition of answer selection and presents the basic context-based and knowledge-based learning framework, which is substantially enhanced in Section~\ref{section33} and Section~\ref{section4}. 
Experimental details and results are discussed in Section~\ref{section5}. 
Finally, we draw a conclusion about the work and provide some possible directions for future works in Section~\ref{section6}.

\section{Related Work}\label{section2}
\subsection{Answer Selection}
The research on answer selection has evolved from early information retrieval (IR) research, which primarily focused on feature engineering~\cite{DBLP:journals/tois/MolinoAL16,DBLP:journals/tois/HuangWNMX19}, syntactic or lexical approach, such as quasi-synchronous grammar~\cite{DBLP:conf/emnlp/WangSM07}, tree edit distance~\cite{DBLP:conf/emnlp/SeverynM13} and lexical semantic model~\cite{DBLP:conf/acl/YihCMP13}. These methods are built on handcrafted features and their performance easily stagnates.

In recent years, deep learning methods make a breakthrough and become the mainstream method to tackle the answer selection task by modeling QA sentences into continuous vector representation without heavy feature engineering. Most of works learn sentence representation via convolutional neural network (CNN)~\cite{DBLP:conf/sigir/SeverynM15} and recurrent neural network such as long-short term memory (LSTM) network~\cite{DBLP:conf/aaai/MuellerT16,DBLP:conf/acl/WangN15}. \cite{DBLP:conf/ijcai/QiuH15} proposes a convolutional neural tensor network (CNTN) architecture to model the interactions between sentences with tensor layers. \cite{DBLP:conf/sigir/TayPLH17} presents Holographic Dual LSTM (HDLSTM) to incorporate holographic representational learning into QA semantic matching. Transfer learning~\cite{DBLP:conf/coling/DengSYLDFL18, DBLP:conf/naacl/ChungLG18} and multi-task learning methods~\cite{DBLP:conf/aaai/DengXLYDFLS19,DBLP:conf/emnlp/ShenSWKLLSZZ18} with deep neural networks are increasingly adopted to introduce information from other domains or tasks to enhance the performance of answer selection.

Additionally, attention mechanism has been proved to be able to significantly improve experimental results on answer selection task by enhancing the interaction between QA pairs~\cite{DBLP:conf/acl/TanSXZ16,DBLP:journals/tacl/YinSXZ16}.
For example, \cite{DBLP:journals/tacl/YinSXZ16} presents an Attention Based Convolutional Neural Network (ABCNN) to incorporate mutual influence between sentences into CNNs. 
\citet{DBLP:conf/acl/TanSXZ16} develops a simple but effective attention mechanism for the purpose of constructing better answer representations according to the input question. 
\citet{DBLP:journals/corr/SantosTXZ16} proposes Attentive Pooling (AP), a two-way attention mechanism for literally triggering semantic interaction between questions and answers. 
\citet{DBLP:conf/acl/WangL016} proposes three kinds of Inner Attention based Recurrent Neural Networks (IARNN), in which attention scheme is added after RNN computation. 
\citet{DBLP:conf/sigir/ChenHHHA17} proposes a positional attention based RNN model to integrate positional information into attentive representation. 

Some latest works employ pre-trained language models, such as ElMo~\cite{DBLP:conf/naacl/PetersNIGCLZ18} or BERT~\cite{DBLP:conf/naacl/DevlinCLT19}, to tackle the answer selection task. \citet{DBLP:conf/cikm/0002DKBJ19} develop a model which adopts ELMo as the pre-trained language model along with some latent clustering and transfer learning techniques. \citet{DBLP:journals/corr/abs-1905-07588} investigate the effectiveness of BERT as the pre-trained language models for answer selection task. Furthermore, \citet{DBLP:conf/emnlp/LaiTBK19} and \citet{DBLP:conf/aaai/XuL20} propose a Gated Self-Attention Memory Network and a Hashing-based Answer Selection model, respectively, to further improve the BERT-based models. \citet{DBLP:conf/aaai/GargVM20} employ transfer learning strategy to leverage additional large-scale QA and natural language inference (NLI) datasets for fine-tuning the BERT encoder. In this work, we also conduct experiments to evaluate whether the external knowledge information is still useful for BERT-based models.

Despite the effectiveness of these studies, they exclusively consider context information rather than real-world background knowledge and hidden information beyond the context. In this paper, we leverage external knowledge to improve the representation learning of deep learning architectures.

\subsection{Integrating External Knowledge}
With the advanced development of knowledge base construction, a large number of large-scale knowledge bases (KBs) such as YAGO~\cite{DBLP:conf/www/SuchanekKW07} and Freebase~\cite{DBLP:conf/sigmod/BollackerEPST08} are available.
As can be seen from many other tasks, it becomes a tendency to exploit external knowledge from KBs to enrich the representational learning in deep learning models. Several efforts have been made on integrating knowledge embeddings trained by knowledge embedding methods~\cite{DBLP:journals/tkde/WangMWG17} to learn a knowledge-aware sentence representation on machine reading~\cite{DBLP:conf/acl/YangM17}, entity typing~\cite{DBLP:conf/aaai/XinL0S18} and relation extraction~\cite{DBLP:conf/aaai/Han0S18}. In this paper, we integrally model contexts and external knowledge into sentence representations, and present the knowledge-aware attention mechanism to explore the interrelation between the knowledge of questions and answers.

On the other hand, existing methods employ the fixed knowledge embeddings trained on external KGs, which are not closely related to the entire sentence. Although researchers~\cite{DBLP:conf/coling/DengSYLDFL18} attempt to alleviate the above issue by learning sentence-level knowledge representations, the use of semantically-fixed initialization of knowledge embeddings (e.g., TransE) can limit the performance of models. To address the limitation, we apply Graph Convolutional Network (GCN)~\cite{DBLP:conf/nips/DefferrardBV16,DBLP:conf/iclr/KipfW17} to learn the contextualized knowledge representation. Our work is inspired by Embeddings from Language Models (ELMo)~\cite{DBLP:conf/naacl/PetersNIGCLZ18} and applications of GCN in NLP ~\cite{DBLP:conf/emnlp/Zhang0M18,DBLP:conf/emnlp/WangLLZ18}.

\subsection{Knowledge Graph Embedding}
Knowledge graph embedding methods aim to assign components of a KG including entities and relations to low-dimensional representations and effectively preserves the KG structure, where KG is regarded as a network or a graph. Therefore, knowledge graph embedding methods can be divided into two categories: network embedding (NE)~\cite{DBLP:journals/corr/abs-1711-08752} and knowledge embedding (KE)~\cite{DBLP:journals/tkde/WangMWG17}. 

NE methods are first developed to embed traditional networks, such as social networks, biological networks, and information networks. Recently, with the rapid growth in KG construction and application, some NE methods, such as Deepwalk~\cite{DBLP:conf/kdd/PerozziAS14} and LINE~\cite{DBLP:conf/www/TangQWZYM15}, are employed to learn the structural representations of KG. 

KE methods directly embed entities and relations into continuous vector representations. There are two types of KE methods: translational distance models and semantic matching models~\cite{DBLP:journals/tkde/WangMWG17}. Translational distance models, such as TransE~\cite{DBLP:conf/nips/BordesUGWY13}, TransD~\cite{DBLP:conf/acl/JiHXL015}, TransH~\cite{DBLP:conf/aaai/WangZFC14}, exploit the relation as a translation and measure the distance between the two entities, while semantic matching models, such as ComplEx~\cite{DBLP:conf/icml/TrouillonWRGB16} and HolE~\cite{DBLP:conf/aaai/NickelRP16}, measure the semantic similarity of entities and relations. In this paper, we study the performance of these knowledge graph embedding methods to enrich the sentence representation on answer selection task.

\begin{table}
	\fontsize{10}{11}\selectfont
	\centering
	\caption{An example for answer selection \label{table:example}}
\begin{tabular}{p{0.1\columnwidth} p{0.42\columnwidth} p{0.36\columnwidth}}
	\toprule
	 & Textual & Knowledge \\
	 \midrule
	 Question & Who established the Nobel Prize awards? & \emph{Nobel\_Prize} \\\\
	 Answer & The Nobel Prize was established in the will of Alfred Nobel, a Swede who invented dynamite and died in 1896. & \emph{Nobel\_Prize}; \emph{the\_will}; \emph{Alfred\_Nobel}; \emph{dynamite} \\
	\bottomrule
\end{tabular}
\end{table}

\section{Background and Basic Framework}\label{section3}
\subsection{Problem Definition}\label{section31}
Given a question $q$ and a set of candidate answers $A = \{a_1,a_2,...,a_n\}$, our model aims to rank the candidate answers by computing a relevancy score $f(q,a)\in [0,1]$ for each question-answer pair. As the example shown in Table~\ref{table:example}, a question or a candidate answer can be represented as a word sequence $W=\{w_1,w_2,...,w_{L_w}\}$. Each question as well as each candidate answer is associated with a piece of knowledge sequence $K=\{k_1,k_2,...,k_{L_k}\}$. 
The knowledge sequence is composed of entity mention sequence which is derived by analyzing the sentence using entity linking~\cite{DBLP:conf/acl/YuYHSXZ17,DBLP:conf/acl/SavenkovA17,DBLP:conf/aaai/DengXLYDFLS19}. The entity linking details are described in Experiment section.

\subsection{Knowledge-aware Neural Network}\label{section32}
\begin{figure}
\centering
\includegraphics[width=0.95\textwidth]{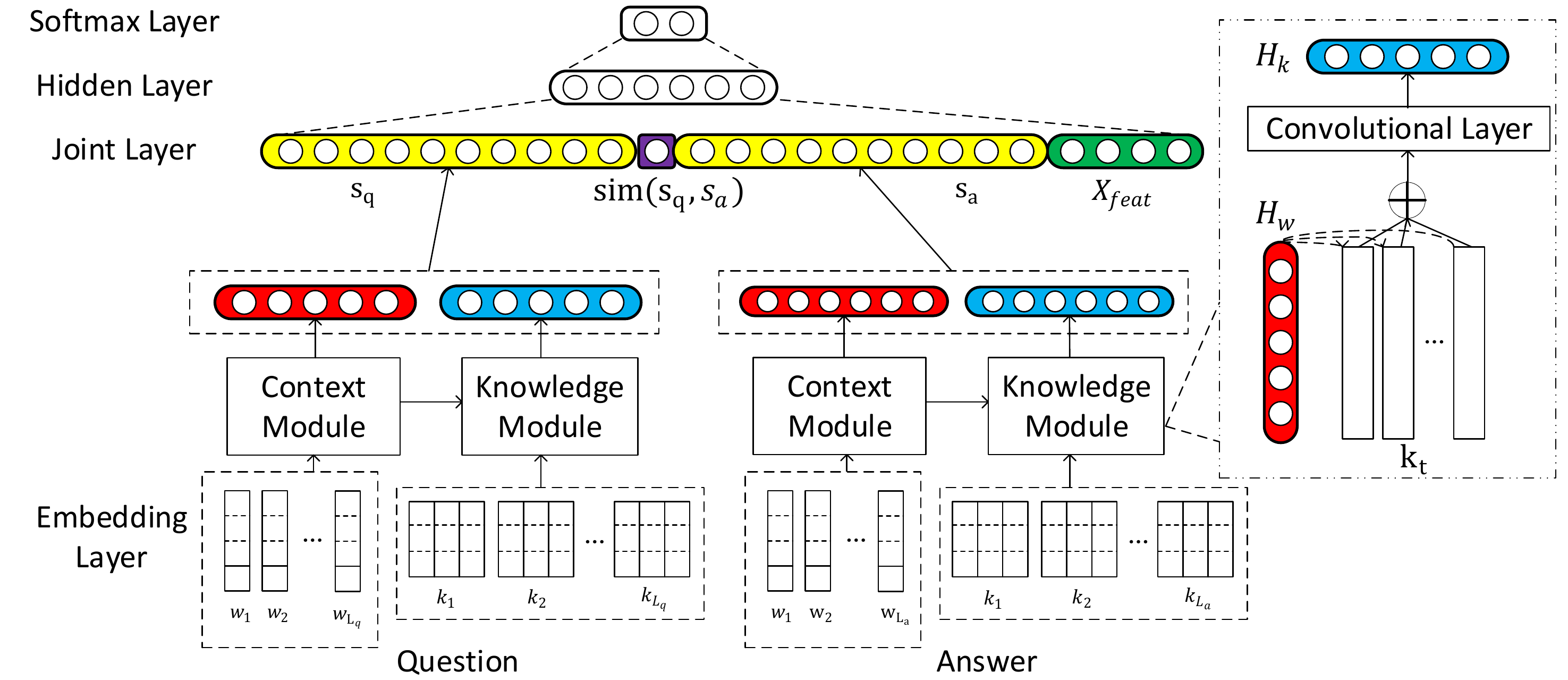}
\caption{Knowledge-aware Neural Network for Answer Selection. Blue, red and yellow matrices denote knowledge-based representations, initial contextual representations and final knowledge-aware sentence representations, respectively.}
\label{fig1}
\end{figure}

In this section, we present a basic framework, called Knowledge-aware Neural Network (KNN), which utilizes external knowledge from knowledge graph (KG) to enrich the sentence representational learning on answer selection. Given a question $q$, the proposed model aims to pick out the correct answers from a set of candidate answers $A=\left\{a_1,\ldots,a_n\right\}$. 
As illustrated in Figure~\ref{fig1}, the overall framework of KNN contains two major components: \emph{Context-based Learning Module} and \emph{Knowledge-based Learning Module}. In Context-based Learning Module, we employ a pair of deep neural networks to learn the initial context-based representations of questions and answers separately (Section~\ref{section321}). In Knowledge-based Learning Module, a context-guided attentive CNN is designed to learn the knowledge-based sentence representation from entities in the sentence (Section~\ref{section322}). After we obtain the context-based and knowledge-based sentence representations, there is a fully connected hidden layer before the final binary classification of joining all the features (Section~\ref{section323}).

\subsubsection{Context-based Learning Module}\label{section321}
Given a question $q$ and a set of candidate answers $A=\left\{a_1,\ldots,a_n\right\}$, we first transform them into vector representations with an embedding layer, and then input these embedding vectors into the Context-based Learning Module. 

Among many deep learning models proposed for answer selection, we adopt the basic Bi-LSTM~\cite{DBLP:conf/acl/WangN15} as the context-based learning model to introduce the overall method. In the experiment section, several popular and typical models, besides Bi-LSTM, are implemented to demonstrate the strong applicability of the proposed method.

In Bi-LSTM, the model not only captures information from past contexts but also has access to future contexts. The Bi-LSTM layer contains two sub-networks for the head-to-tail and the tail-to-head context respectively. The output at the time step $t$ is represented by $h_t=[\overrightarrow{h_t}:\overleftarrow{h_t}]$, in which $\overrightarrow{h_t}$ is the output of the forward network and $\overleftarrow{h_t}$ is the output of the backward network. Given the question $q$ and the answer $a$, we generate the initial contextual sentence representation $H_w\in\mathbb{R}^{L\times{d_h}}$ for both the question and the answer, where $L$ and $d_h$ are the length of sentences and the dimension of $h_t$:
\begin{equation}
   Q_w = \text{Bi-LSTM}(W_q),\quad
   A_w = \text{Bi-LSTM}(W_a).
\end{equation}

\subsubsection{Knowledge-based Learning Module}\label{section322}
Knowledge-based learning module is designed to learn knowledge-based sentence representations from discrete candidate entity embeddings with the guidance of contextual information. We perform entity mention detection by n-gram matching and provide a set of top-$K$ entity candidates from KG for each entity mention in the sentence. Due to the ambiguity of the entity, e.g., ``Boston'' can refer to a city or a person, we design a context-guided attention mechanism to learn the knowledge representation of each entity mention in the sentence by congregating the embeddings of the corresponding candidate entities in the KG. The contextual sentence representations of questions and answers are learned by Bi-LSTM layers, while the embeddings of entities in KG are pretrained by TransE~\cite{DBLP:conf/nips/BordesUGWY13}. Formally, we present candidate entities for the entity mention at the time step $t$ as $k_t=\left\{e_1,e_2,\ldots,e_K\right\}\in\mathbb{R}^{K\times{d_e}}$, where $d_e$ is the dimension of the entity embedding in KG. Note that $k_t$ will be padded with zero embeddings if there is no candidate entity at the time step $t$. Then, the final context-guided embedding for the word at the time step $t$ (in accord with $t$ in Section~\ref{section321}) is given by
\begin{equation}
 m_t=\tanh\left(W_{em}k_t+W_{hm}H_w\right),
\end{equation}
\begin{equation}
 \alpha_t=\frac{exp\left(w_{m}^\intercal m_{t_i}\right)}{\sum_{m_{t_j}\in m_t} exp\left(w_{m}^\intercal m_{t_j}\right)} ,
\end{equation}
\begin{equation}
 \widetilde {k_t}=\sum_{e_{t_i}\in k_t} \alpha_{t_i} e_{t_i},
\end{equation}
where $W_{em}$, $W_{hm}$ and $w_{m}$ are parameter matrices to be learned. $m\left(t\right)$ is a context-guided knowledge vector, and $\alpha_{t}$ denotes the context-guided attention weight that is applied over each candidate entity embedding $e_{t_i}$. The contextual sentence representations $H_w$ of questions and answers are learned by the base model, while the embeddings of entities in KG are learned by knowledge embedding methods (The details will be presented in Section~\ref{section5}).

This procedure produces a context-guided representation for each entity mention in the sentence. Then we apply an CNN layer to learn a higher level knowledge-based sentence representation. The input of the CNN layer are the attentive knowledge embeddings $\widetilde K = \left\{\widetilde {k_1},\widetilde {k_2},\ldots,\widetilde {k_L}\right\} \in\mathbb{R}^{L\times{d_e}}$.

In the convolution layer, a filter of size $n$ slides over the input embedding matrix to capture the local n-gram information, which is useful to extract the entity features since an entity is likely to be a phrase. Each move computes a hidden layer vector as
\begin{equation}
 x_i = [\widetilde k_{i-\frac{n-1}{2}},\ldots,\widetilde k_i,\ldots,\widetilde k_{i+\frac{n-1}{2}}],
\end{equation}
\begin{equation}
 h_i = \tanh\left(Wx_i+b\right),
\end{equation}
where $W$ and $b$ are the convolution kernel and the bias vector to be learned, respectively. Then we employ max-pooling over the hidden layer vectors $h_1,\ldots,h_n$ to generate the final output vector $y$:
\begin{equation}
 y_j = max\left\{{h_1}_j,\ldots,{h_n}_j\right\},
\end{equation}
where $y_j$ and ${h_i}_j$ are the $j$-th value of the output vector $y$ and the hidden vector $h_i$, respectively.

Due to the uncertainty of the length of entities, we exploit several filters of various sizes to obtain n-gram features $\left\{y^{\left(1\right)},y^{\left(2\right)},\ldots,y^{\left(n\right)}\right\}$, where $y^{\left(i\right)}$ denotes the output vector obtained by the $i$-th filter. We pass these output vectors through a fully-connected layer to get the final knowledge-based sentence embedding $H_k\in\mathbb{R}^{L\times{d_f}}$, where $d_f$ is the total filter sizes of CNN and $L$ is the length of the sentence. As for the question $q$ and the answer $a$, we generate their knowledge-based sentence representations $Q_k$ and $A_k$ as:
\begin{equation}
Q_k = [y^{\left(1\right)}_q,y^{\left(2\right)}_q,\ldots,y^{\left(n\right)}_q],\quad
A_k = [y^{\left(1\right)}_a,y^{\left(2\right)}_a,\ldots,y^{\left(n\right)}_a].
\end{equation}

\subsubsection{Hidden Layer and Softmax Layer}\label{section323}
Following the strategies in~\cite{DBLP:conf/sigir/SeverynM15,DBLP:conf/sigir/TayPLH17}, additional features are exploited in our overall architecture. First, we compute the bilinear similarity score between final attentive QA vectors:
\begin{equation}
 sim\left(s_q,s_a\right)=s_q^\intercal Ws_a,
\end{equation}
where $W\in\mathbb{R}^{L\times{L}}$ is a similarity matrix to be learned. Besides, the same word overlap features $X_{feat}\in\mathbb{R}^4$ are incorporated into our model, which can be referred to~\cite{DBLP:conf/sigir/SeverynM15,DBLP:conf/sigir/TayPLH17}. Thus, the input of the hidden layer is a vector $\left[s_q,sim(s_q,s_a),s_a,X_{feat}\right]$, and its output then goes through a softmax layer for binary classification. The overall model is trained to minimize the cross-entropy loss function:
\begin{equation}
 L=-\sum_{i=1}^N\left[y_i\log{p_i}+\left(1-y_i\right)\log{\left(1-p_i\right)}\right]+\lambda{\Vert\theta\Vert}_2^2,
\end{equation}
where $p$ is the output of the softmax layer. $\theta$ contains all the parameters
of the network and $\lambda{\Vert\theta\Vert}_2^2$
is the L2 regularization.

\section{Knowledge-aware Attention Mechanism}\label{section33}
As for a question vector $q$ and an answer vector $a$, we obtain context-based sentence representations $H_w$ for the question and the answer from the context module and the knowledge-based representations $H_k$ from the knowledge module. Afterwards, we propose two kinds of knowledge-aware attention mechanism to learn the final knowledge-aware attentive sentence representation, namely \textit{Knowledge-aware Self-Attention Mechanism} and \textit{Knowledge-aware Co-Attention Mechanism}. Knowledge-aware attention mechanism is an approach that enables the original model to be aware of the background information and the relation of knowledge beyond the text. We refer the Knowledge-aware Neural Network (KNN) equipped with Knowledge-aware Attention Mechanism as Knowledge-aware Attentive Neural Network (KANN). Note that the additive and dot product attention computations can be easily replaced by multi-head attention computation~\cite{DBLP:conf/nips/VaswaniSPUJGKP17}. For simplification, vanilla attention computations are adopted.

\subsection{Knowledge-aware Self-Attention Mechanism}\label{section331}

\begin{figure}
\centering
\includegraphics[width=0.8\textwidth]{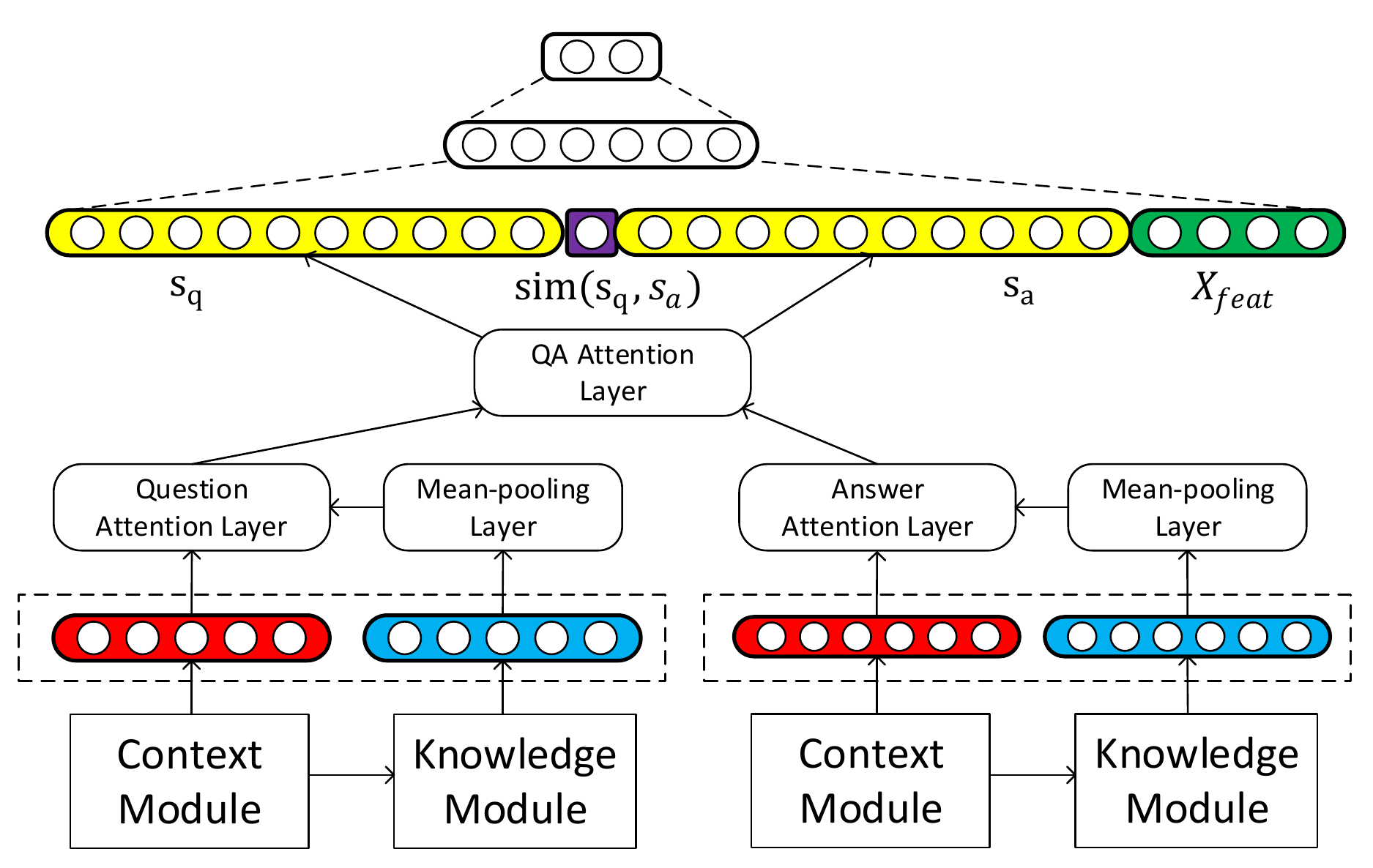}
\caption{Knowledge-aware Attentive Neural Network with Knowledge-aware Self-Attention Mechanism.}
\label{fig2}
\end{figure}

Knowledge-aware self-attention mechanism exploits knowledge-based sentence representations learned from the knowledge module to attend knowledge information in context-based sentence representations of the question and the answer themselves, respectively. 
Figure~\ref{fig2} illustrates the architecture of Knowledge-aware Attentive Neural Network (KANN) with Knowledge-aware Self-Attention Mechanism.

We exploit the overall knowledge information as the attention source over all the elements in the sentence. Thus, we apply mean pooling over the knowledge-based representations $Q_k$ and $A_k$ from knowledge module:
\begin{equation}
 o_{q_k} =\text{Mean-Pooling}\left(Q_k\right), \quad
 o_{a_k} =\text{Mean-Pooling}\left(A_k\right).
\end{equation}

Conceptually, the attention mechanism takes into consideration the knowledge information,  which is expected to capture the background information in the question and the answer:
\begin{gather}
 \alpha_q^{k}=\text{softmax}(w^\intercal\text{tanh}(W_{1}o_{q_k}+W_{2}Q_w)),\\
 \alpha_a^{k}=\text{softmax}(w^\intercal\text{tanh}(W_{1}o_{a_k}+W_{2}A_w)).
\end{gather}
where $w\in\mathbb{R}^{d_h}$, $W_{1}, W_{2}\in\mathbb{R}^{d_h\times{d_h}}$ are attention parameters to be learned; $\alpha_q^{k}$ and $\alpha_a^{k}$ are the knowledge-aware self-attention weights for the question and the answer respectively. The knowledge-aware attentive sentence representations will be:
\begin{equation}
O_q = [Q_w:Q_k]^\intercal\alpha_q^{k}, \quad
O_a = [A_w:A_k]^\intercal\alpha_a^{k}.
\end{equation}

In the QA attention layer, we employ a two-way attention mechanism to generate the attention between final question and answer representations:
\begin{gather}
 M_{qa}=\text{tanh}\left(O_q^\intercal U_{qa}O_a\right),\\
 \alpha_q^{o} = \text{softmax}(\text{Max-Pooling}(M_{qa})),\\
 \alpha_a^{o} =\text{softmax}(\text{Max-Pooling}({M_{qa}}^\intercal))\label{eq1},
\end{gather}
where $U_{qa}\in\mathbb{R}^{d_{h}\times{d_{h}}}$ is the attention parameter matrix to be learned; $\alpha_q^{o}$ and $\alpha_a^{o}$ are the attention weights for the question and the answer, respectively. The final sentence representations for the question and the answer will be:
\begin{equation}
s_q = O_q\alpha_q^{o}, \quad s_a = O_a\alpha_a^{o}.
\end{equation}

\subsection{Knowledge-aware Co-Attention Mechanism}\label{section332}

\begin{figure}
\centering
\includegraphics[width=0.8\textwidth]{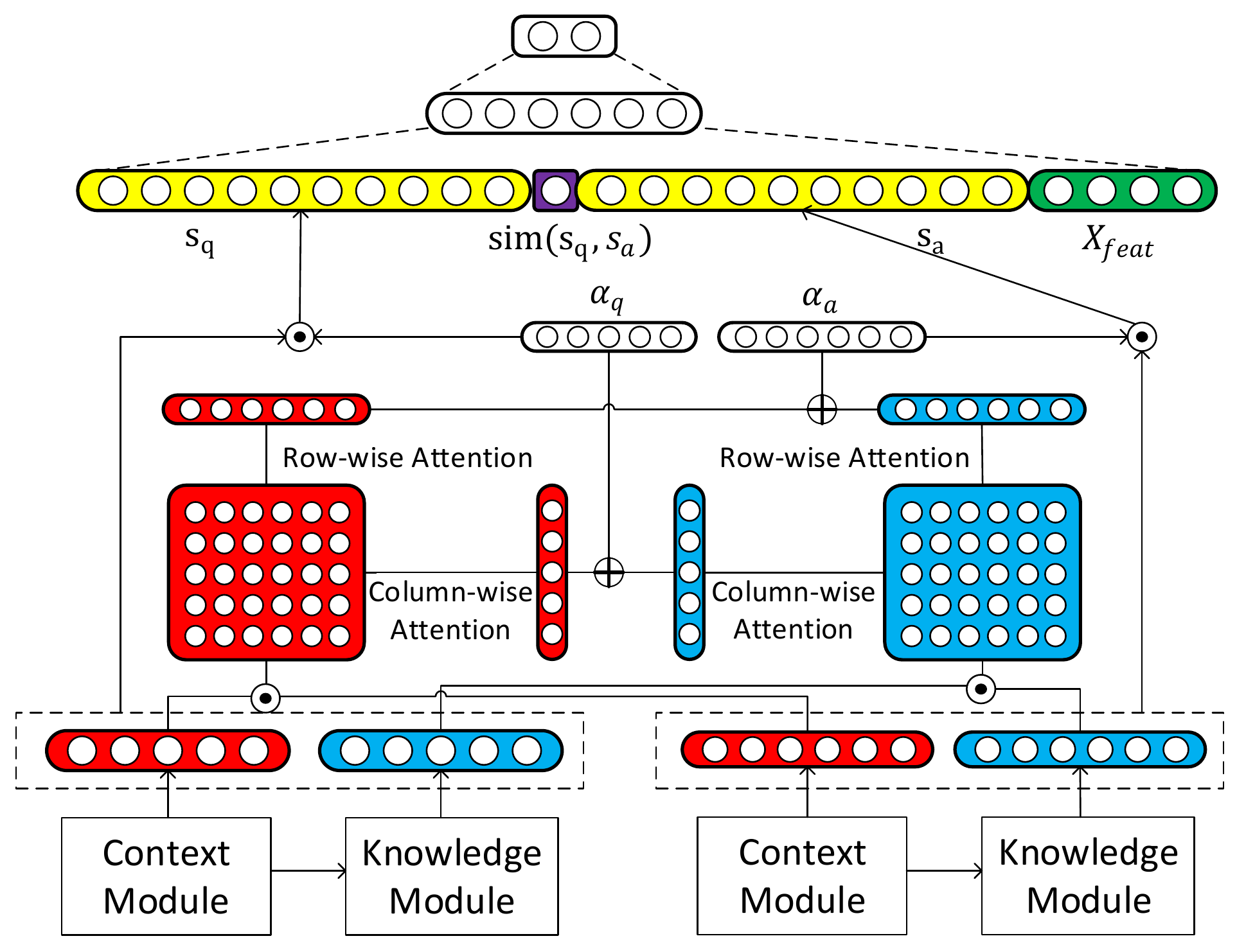}
\caption{Knowledge-aware Attentive Neural Network with Knowledge-aware Co-Attention Mechanism.}
\label{fig3}
\end{figure}

Knowledge-aware co-attention mechanism integrates context-based and knowledge-based sentence representations to learn co-attention weights between questions and answers, which can gather not only important textual information but also useful knowledge information between QA sentences.

For both question and answer sentences, there are two different sentence-level representation vectors. $Q_w$ and $A_w$ are learned from word embeddings, while $Q_k$ and $A_k$ are derived from knowledge module. These two kinds of sentence vectors are input into knowledge-aware co-attention mechanism. Figure~\ref{fig3} illustrates the architecture of KANN with Knowledge-aware Co-Attention Mechanism.

As illustrated in Figure~\ref{fig3}, we first compute two attention matrices $M_w$ and $M_k$:
\begin{equation}
 M_w=tanh\left(Q_w^\intercal U_wA_w\right), \quad
 M_k=tanh\left(Q_k^\intercal U_kA_k\right),
\end{equation}
where $U_w\in\mathbb{R}^{d_h\times{d_h}}$ and $U_k\in\mathbb{R}^{d_f\times{d_f}}$ are parameter matrices to be learned. 

Then column-wise and row-wise max-pooling are applied on $M_w$ to generate context-based attention vectors  for the question and the answer separately, while we conduct the same operation over $M_k$ for knowledge-based attention vectors. In order to incorporate the knowledge-aware influence of the question words into answers' attentive representations and vice versa, we merge these two attention vectors to obtain the final knowledge-aware attention vectors, $\alpha_q$ and $\alpha_a$:
\begin{equation}
\small
 \alpha_q\propto\left(\text{softmax}\left(\text{Max-Pooling}_{1<l<L_q}M_w\right)+\text{softmax}\left(\text{Max-Pooling}_{1<l<L_q}M_k\right)\right),
\end{equation}
\begin{equation}
\small
 \alpha_a\propto\left(\text{softmax}\left(\text{Max-Pooling}_{1<l<L_a}M_w^\intercal\right)+\text{softmax}\left(\text{Max-Pooling}_{1<l<L_a}M_k^\intercal\right)\right).\label{eq2}
\end{equation}

We conduct dot product between the attention vectors and the overall sentence vectors to form the final knowledge-aware attentive sentence representations of question $q$  (i.e., $s_q$) and answer $a$ (i.e., $s_a$):
\begin{gather}
 s_q = [Q_w:Q_k]^\intercal\alpha_q, \quad
 s_a = [A_w:A_k]^\intercal\alpha_a,
\end{gather}
where [:] is the concatenation operation.

Then the knowledge-aware attentive representations $s_q$ and $s_a$ are input into the joint layer as KNN to measure the correlation degree between the question and the answer.

\begin{figure}
	\centering
	\includegraphics[width=\textwidth]{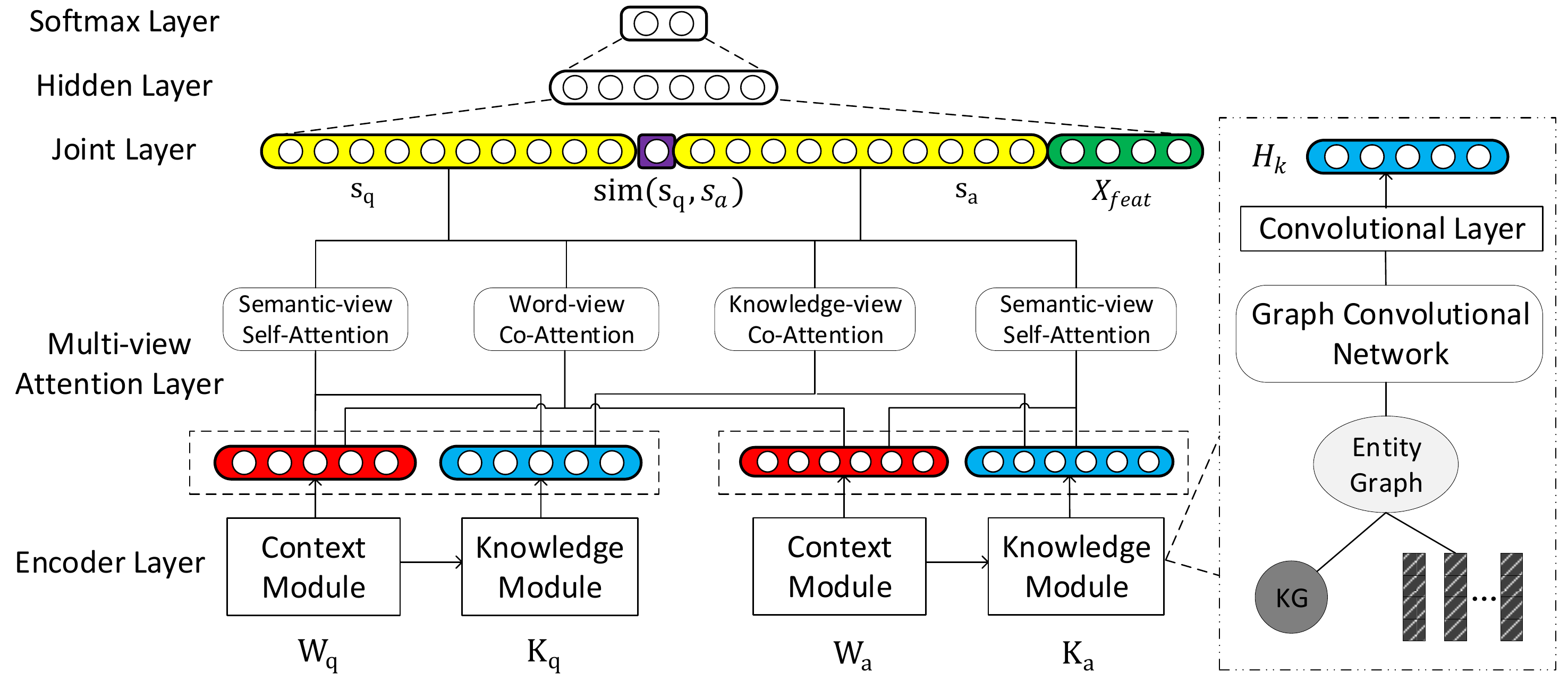}
	\caption{Contextualized Knowledge-aware Attentive Neural Network \label{figure:model}}
\end{figure}

\section{Contextualized Knowledge-aware Attentive Neural Network}\label{section4}
In order to dynamically and adaptively encode the knowledge information as well as comprehensively model the interactions between context and knowledge information, we further develop a Contextualized Knowledge-aware Attentive Neural Network (CKANN) model, which is based on the KNN framework and knowledge-aware attention mechanism. An overview of CKANN is illustrated in Figure~\ref{figure:model}. 

Concretely, we first employ a pair of Bidirectional Long Short-Term Memory (Bi-LSTM) to learn the context-based sentence representations of questions and answers respectively, which is the same as the KNN framework. Differently, we generate the Entity Graph with external Knowledge Graph (KG) mapping and learn knowledge-based sentence representations by applying Graph Convolutional Networks (GCN) to capture certain structure information from the Entity Graph, instead of explicitly using the fixed pre-trained knowledge embeddings. Then we develop a multi-view attention mechanism, combining the Knowledge-aware Self-Attention and Knowledge-aware Co-Attention, to comprehensively fetch the semantic relationship of words and knowledge between each QA pair. Finally, a fully connected hidden layer before the final output layer is used to join features of QA pairs as the KNN framework. In the rest of this section, we elaborate CKANN model in detail, including Contextualized Knowledge-based Learning Module (Section~\ref{section41}) and Multi-view Attention (Section~\ref{section42}). Since the context-based learning module and the output layer are in accordance with the KNN framework, we omit these descriptions.

\subsection{Contextualized Knowledge-based Learning Module}\label{section41}
In the \textbf{Knowledge-based Learning Module} from KNN (Section~\ref{section322}), the knowledge-based sentence representation is directly learned from the pre-trained knowledge embeddings. As one may expect, the same knowledge appearing in different sentences is likely to be diverse in their information. Besides, structure information is supposed to be an essential advantage of knowledge graph. Thus, we enhance the knowledge-based learning module into a contextualized knowledge-based learning module by dynamically taking into account the graph structure information.

\subsubsection{Entity Graph Generation}\label{section411}
In specific, we perform entity mention detection by entity linking~\cite{DBLP:conf/acl/SavenkovA17} and construct a knowledge sequence $K=\{k_1,k_2,...,k_{L_k}\}$ for each question and each candidate answer. Then we build the Entity Graphs as follows: (i) convert knowledge $k$ into entity nodes (denoted as \textbf{original entity nodes}) via mapping them into external KG, and reserve the edges among them; (ii) add one-hop neighbors (denoted as \textbf{new added entity nodes}) of the original entity nodes and reserve the edges from KG to introduce some relevant background information and semantic relationship beyond words; (iii) add edges to the contiguous sequence of $p$ entities that appear in a given sentence to enrich their interrelations, denoted as \textbf{newly added edges}. The reserved edges from KG are denoted as \textbf{original edges}. The generated Entity Graph is an undirected graph with at most one edge between every entity node pair. 

\begin{figure}
	\centering
	\includegraphics[width=\textwidth]{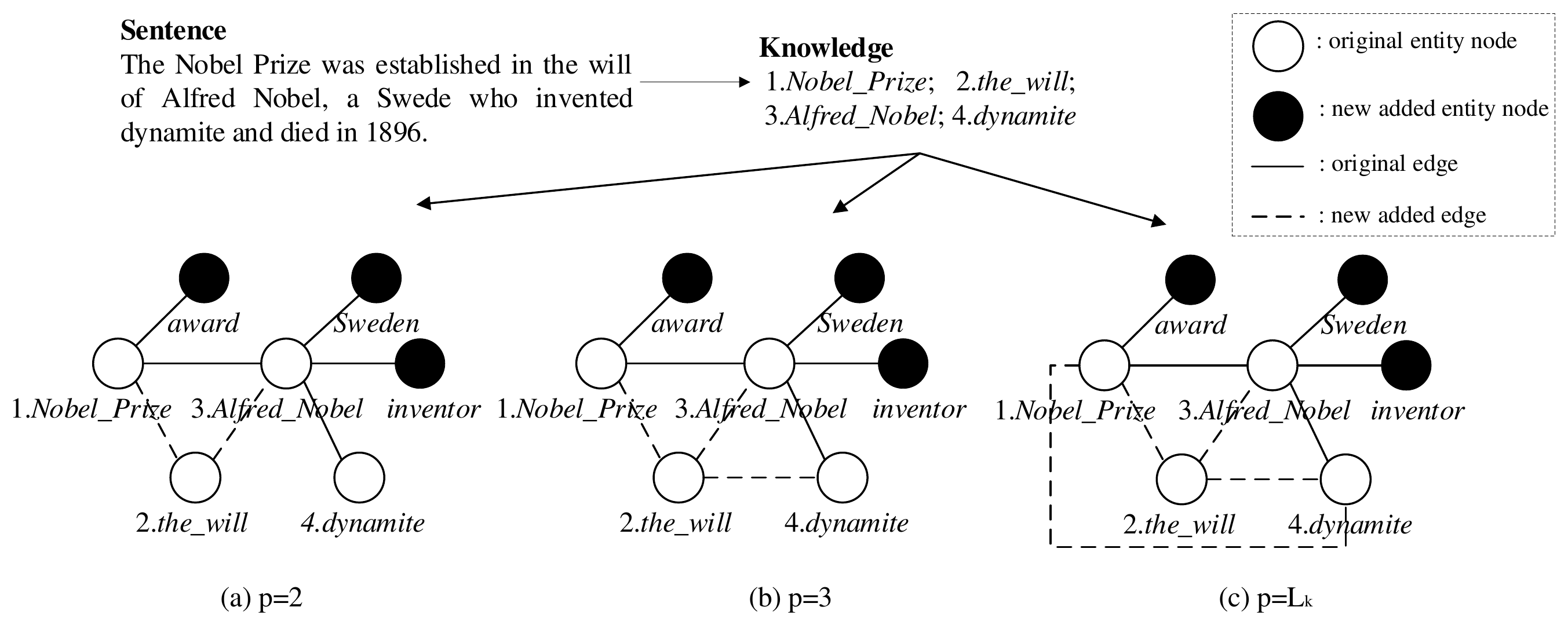}
	\caption{Entity Graphs with different $p$ values. The white entity nodes are original entity nodes, obtained by mapping the knowledge sequence into KGs. The black nodes are their one-hop neighbors. The edges in black denote the reserved edges from KGs. Besides, we add edges to enrich the relationship among nodes, which are indicated by dotted lines. \label{fighre:entity graph}}
\end{figure}

Figure~\ref{fighre:entity graph} shows an example of the Entity Graph. Entities in a given sentence are numbered sequentially to facilitate the reader's understanding. In our study, we build three types of Entity Graph with different $p$ values:
\begin{itemize}
	\item $p=2$. We add an edge between each original entity node pair that appears sequentially in a given sentence, emphasizing the relationship of adjacent entity nodes. For example, we add an edge between 1.\emph{Nobel\_Prize} and 2.\emph{the\_will}, as shown in Figure~\ref{fighre:entity graph}(a).
	
	\item $p=3$. 
	Considering that neighboring entity nodes may be semantically related, we choose the three original entity nodes that appear sequentially and connect each of them with edges. As shown in Figure~\ref{fighre:entity graph}(b), given an entity node sequence \{2.\emph{the\_will}, 3.\emph{Alfred\_Nobel}, 4.\emph{dynamite}\}, we add edges between 2.\emph{the\_will} and 3.\emph{Alfred\_Nobel}, 3.\emph{Alfred\_Nobel} and 4.\emph{dynamite}, as well as 2.\emph{the\_will} and 4.\emph{dynamite}.
	
	\item $p=L_k$. A complete graph is generated among all the original entity nodes that appear in a given question/answer, i.e., each pair of original entity nodes are connected by an edge (see Figure~\ref{fighre:entity graph}(c)).
\end{itemize}

\subsubsection{Graph Convolutional Network}
We adopt TransE~\cite{DBLP:conf/nips/BordesUGWY13} as the knowledge embedding method to generate initial embeddings for all nodes in the Entity Graph. Given a knowledge sequence $K=\{k_1,k_2,...,k_{L_k}\}$, we transform it into the embedding representation as $E_k=\{e_{k_1},e_{k_2},...,e_{k_{L_k}}\} \in \mathbb{R}^{d_k \times L_k}$, where $L_k$ is the knowledge length and $d_k$ is the embedding dimension.

However, knowledge generally does not exist independently. Knowledge representations should be adaptive based on the entire input sentence, instead of assigning fixed pre-trained embeddings. The relationships among the knowledge in the Entity Graph make the semantic representations of knowledge vary across contexts and beyond the fixed embeddings. To directly address this, we learn the contextualized knowledge representation by applying Graph Convolutional Network (GCN)~\cite{DBLP:conf/nips/DefferrardBV16} on the Entity Graph.

We use an one-localized convolution to define our Graph Convolutional Network on the Entity Graph, i.e., the contextualized knowledge representation is only affected by entity nodes that are maximum one step away from it (one-hop neighbor entities in the Entity Graph), which is similar to \cite{DBLP:conf/iclr/KipfW17}. Formally, the propagation rule of GCN is given as:
\begin{equation}
H_{out} = \sigma(D^{-\frac{1}{2}}AD^{-\frac{1}{2}}H_{in}W),
\end{equation}
where $\sigma(\cdot)$ denotes an activation function, $W \in \mathbb{R}^{d_k \times d_k} $ is the weight matrix to be learned and $d_k$ is the dimension of entity embedding. $H_{in}$ and $H_{out}$ denote the input and output of GCN. Here, $A \in \mathbb{R}^{L_k \times L_k}$ and $D \in \mathbb{R}^{L_k \times L_k}$ is the adjacency matrix and degree matrix of the Entity Graph with added self-connections, respectively:
\begin{align}
A_{ij} & := \begin{cases}
1 &\text{if } i=j \text{ or } {v_i} \text{ is adjacent to } v_j;\\
0 &\text{otherwise.}
\end{cases} \\
D_{ij} & := \begin{cases}
\sum_{k}A_{ik} &\text{if } i=j ;\\
0 &\text{otherwise.}
\end{cases} 
\end{align}
where $v_i$ and $v_j$ denote the entity nodes in the Entity Graph. We feed the knowledge embedding representation into GCN and obtain the high-quality contextualized knowledge representation:
\begin{equation}
\hat{E}_k = \text{GCN}(E_k).
\end{equation} 

Finally, we apply the mean-pooling operation to summarize and smooth three kinds of knowledge representation, which are obtained via applying GCN on three types of Entity Graphs with different $p$ values $(p=2,3,L_k)$. Thus, we get the final contextualized knowledge representation:
\begin{equation}
\widetilde{E}_k = \text{Mean-Pooling}(\hat{E}_k^{(2)},\hat{E}_k^{(3)},\hat{E}_k^{(L_k)}).
\end{equation} 

\subsubsection{Convolutional Neural Network} 
We adopt a CNN to obtain the knowledge-based sentence representations. A filter of size $n$ is applied to slide through the knowledge representation matrix to capture the local $n$-gram features, which is formally given as:
\begin{gather}
\widetilde{E}_{k_t} = [\widetilde{e}_{k_t},\widetilde{e}_{k_{t+1}},...,\widetilde{e}_{k_{t+n-1}}],\\
h_t = \sigma(\widetilde{E}_{k_t}*W_c + b_c), 
\end{gather}
where $*$ is convolution operator, $W_c\in \mathbb{R}^{d_k\times n}$ and $b_c$ are the convolution kernel matrix and the bias vector to be learned, respectively.

Further, a couple of filters of various sizes are employed to obtain several output feature maps $\{H_{k_1}, H_{k_2},...,H_{k_n}\}$, where $H_{k_i}$ is the output vector obtained by the $i$-th filter. Finally, we get the knowledge-based sentence representation $H_k \in \mathbb{R}^{d_h\times L_k}$ for each question/answer sentence by concatenating the output feature maps:
\begin{equation}
H_k = [H_{k_1}^{\top}: H_{k_2}^{\top}:...:H_{k_{d_h}}^{\top}],
\end{equation}
where $L_k$ is the length of knowledge sequence and $d_h$ is the total filter number of CNN, which is set to be the same as the hidden dimension of Bi-LSTM in \textbf{Context-based Learning Module} (Section~\ref{section321}).

\subsection{Multi-view Attention}\label{section42}
We develop a multi-view attention mechanism from the basic \textbf{Knowledge-aware Attention Mechanism} (Section~\ref{section33}) to capture the semantic relationship of words and knowledge between QA pairs, which adaptively decides the importance of different words and knowledge in questions and answers. Specifically, we compute three views of attention: word-view co-attention, knowledge-view co-attention, and semantic-view self-attention. 

\subsubsection{Word-View \& Knowledge-View Co-Attention}
We apply word-view and knowledge-view co-attention mechanism to calculate the attention between each question-answer pair, which enables question-answer pairs to be aware of the semantic potential relationship between each other. Note that the combination of these two views of co-attention is equivalent to \textbf{Knowledge-aware Co-attention} (Section~\ref{section331}).

For words of question-answer pair, we have learned context-based sentence representation $Q_w$ and $A_w$ from \textbf{Context-based Learning Module}. Then we compute the word-view co-attention matrix $M_w$:
\begin{equation}
M_w = Q_w^{\top}U_w A_w,
\end{equation}
where $U_w \in \mathbb{R}^{d_h \times d_h}$ is the parameter matrix to be learned. We apply column-wise and row-wise max-pooling on $M_w$ to generate the word view attention weights $\alpha_w^Q$ and $\alpha_w^A$ for question and answer separately:
\begin{gather}
\alpha_w^Q = \text{softmax}\left(\text{Max-Pooling}(M_w)\right), \\
\alpha_w^A = \text{softmax}\left(\text{Max-Pooling}(M_w^{\top})\right). 
\end{gather}

We adopt the similar operation on knowledge-based sentence representation of each QA pair to obtain the knowledge view attention weights $\alpha_k^Q$ and $\alpha_k^A$, which can be formally given as:
\begin{gather}
M_k = Q_k^{\top}U_kA_k,\\
\alpha_k^Q = \text{softmax}\left(\text{Max-Pooling}(M_k)\right), \\
\alpha_k^A = \text{softmax}\left(\text{Max-Pooling}(M_k^{\top})\right), 
\end{gather}
where $U_k \in \mathbb{R}^{d_k \times d_k}$ is the parameter matrix to be learned. Here, $Q_k$ and $A_k$ denote the knowledge-based sentence representation of the question and the answer respectively.

\subsubsection{Semantic-View Self-Attention} 
The intuition of semantic-view attention is to perceive the importance of different items (words or knowledge) based on the semantic correlations between words and knowledge. Also, the semantic-view self attention is derived from the \textbf{Knowledge-aware Self-Attention Mechanism} (Section~\ref{section332}).

Given the context-based and knowledge-based sentence representation, we first apply a max-pooling operation on knowledge-based sentence representation, which can be regarded as a selector to preserve the main features from knowledge:
\begin{gather}
\hat{Q}_k = \text{Mean-Pooling}(Q_k),\\
\hat{A}_k = \text{Mean-Pooling}(A_k).
\end{gather}

Then we concatenate the results and context-based sentence representation to generate the semantic-aware context-based representation $\widetilde{Q}_w$ and $\widetilde{A}_w$. The semantic view attention weights for the word sequence is given by:
\begin{gather}
\widetilde{Q}_w = [Q_w:\hat{Q}_k],\quad \beta_w^Q = u_w^Q \tanh(W_w^Q \widetilde{Q}_w),\\
\widetilde{A}_w = [A_w:\hat{A}_k],\quad \beta_w^A = u_w^A \tanh(W_w^A \widetilde{A}_w),
\end{gather}where $[:]$ denotes the concatenation operation, $W_w^Q, W_w^A\in \mathbb{R}^{d_h\times (2*d_h)}$ and $u_w^Q, u_w^A\in \mathbb{R}^{d_h}$ are parameters to be learned. 
Thus, the semantic-view attention mechanism takes into consideration the semantic correlations between words and knowledge.

On the other hand, we feed the context-based sentence representation into the max-pooling layer, and obtain the semantic-aware knowledge representation $\widetilde{Q}_k$ and $\widetilde{A}_k$ via concatenation:
\begin{gather}
\hat{Q}_w = \text{Mean-Pooling}(Q_w),\quad \widetilde{Q}_k = [Q_k:\hat{Q}_w],\\
\hat{A}_w = \text{Mean-Pooling}(A_w),\quad \widetilde{A}_k = [A_k:\hat{A}_w].
\end{gather}
We adopt the similar operation and obtain the semantic view attention weights for knowledge:
\begin{gather}
\beta_k^Q = u_k^Q \tanh(W_k^Q \widetilde{Q}_k),\\
\beta_k^A = u_k^A \tanh(W_k^A \widetilde{A}_k),
\end{gather}
where $W_k^Q, W_k^A\in \mathbb{R}^{d_h\times (2*d_h)}$ and $u_k^Q, u_k^A\in \mathbb{R}^{d_h}$ are parameters to be learned.

\subsubsection{Multi-View Attention Representation}
We define the multi-view attention fusion from word, knowledge and semantic views as:
\begin{gather}
\gamma_w^Q = \text{softmax}(\alpha_w^Q + \beta_w^Q),\\
\gamma_k^Q = \text{softmax}(\alpha_k^Q + \beta_k^Q),\\
\gamma_w^A = \text{softmax}(\alpha_w^A + \beta_w^A),\\
\gamma_k^A = \text{softmax}(\alpha_k^A + \beta_k^A).
\end{gather}

Thus, the final attentive representations of question and answer under multi-view attention are given by
\begin{gather}
s_w^Q = Q_w \gamma_w^Q,\quad s_w^A = A_w \gamma_w^A;\quad\\ 
s_k^Q = Q_k \gamma_k^Q,\quad s_k^A = A_k \gamma_k^A;\quad \\
s_q = [s_w^Q:s_k^Q],\quad s_a = [s_w^A:s_k^A].
\end{gather} 

Finally, the knowledge-aware attentive representations $s_q$ and $s_a$ are fed into the joint layer as KNN to measure the correlation degree between the question and the answer.
\section{Experiment}\label{section5}
We conduct experiments to validate the effectiveness of the proposed method on answer selection from the following aspects: (1) We compare the proposed method with the state-of-the-art methods on four benchmark answer selection datasets, including TREC QA, WikiQA, InsuranceQA and Yahoo QA (Section~\ref{subsection52}); (2) We conduct ablation test to testify the effect of different components and additional features in the proposed method (Section~\ref{subsection56}); (3) We implement several context-based deep learning models into the proposed knowledge-aware neural network framework to show the applicability and universality of the proposed method (Section~\ref{subsection53}); (4) We compare the performance of various popular knowledge embedding methods on answer selection (Section~\ref{subsection54}); (5) We analyze  the  performance  of  our  model  with  respect  to the  completeness of the knowledge graph (Section~\ref{subsection55}); (6) We visualize the attention weights to demonstrate the effectiveness of the knowledge-aware attention mechanism (Section~\ref{subsection57}).
\subsection{Experimental Setup}\label{subsection51}
\subsubsection{Datasets and Metrics}\label{subsubsection511}
We evaluate our method on four widely-used QA benchmark datasets: TREC QA, WikiQA, InsuranceQA and Yahoo QA, in which the first two datasets are concerned about factoid questions, while the other two consist of non-factoid questions. The statistics of these datasets are described in Table~\ref{datasets}.

\begin{table}
\centering
  \caption{Summary statistics of datasets.}
  \begin{tabular}{cccc}
    \toprule
	Dataset &\multirow{2}{*}{Type}& \multirow{2}{*}{\#Question} & \multirow{2}{*}{\#QA Pairs} \\
    (train/dev/test) &&& \\
    \midrule
   TREC QA(original) &Factoid& 1,229/82/100 & 53,417/1,148/1,517\\
   TREC QA(cleaned) &Factoid& 1,160/65/68 & 53,313/1,117/1,442 \\
    WikiQA&Factoid& 873/126/243 &8,627/1,130/2,351 \\
    InsuranceQA&Non-factoid& 12,887/1,000/1,800x2 & 37.1K/500K/900Kx2\\
    Yahoo QA &Non-factoid& 50,098/6,289/6,283 & 253K/31K/31K\\
  \bottomrule
\label{datasets}
\end{tabular}
\end{table}

\textbf{TREC QA}. Original TREC QA dataset, collected from TREC QA track 8-13 data~\cite{DBLP:conf/emnlp/WangSM07}, is a widely-adopted benchmark for factoid question answering. Besides, there is a cleaned version of TREC QA dataset, which removes questions that have only positive or negative answers or no answer. As for TREC QA, following previous work \cite{DBLP:conf/emnlp/WangSM07,DBLP:conf/emnlp/YangYM15}, the mean average precision (MAP) and mean reciprocal rank (MRR) are adopted as our evaluation metrics.

\textbf{WikiQA}. WikiQA dataset is an open-domain factoid answer selection benchmark, in which the standard pre-processing steps as~\cite{DBLP:conf/emnlp/YangYM15} is employed to extract questions with correct answers. Same as TREC QA, MAP and MRR are also adopted to evaluate the performance on WikiQA.

\textbf{InsuranceQA}. A community-based QA dataset, proposed by ~\cite{DBLP:conf/asru/FengXGWZ15}, contains QA pairs from the insurance domain and is split into a training set, a development set, and two test sets. For the development and test sets, the InsuranceQA also includes an answer pool of 500 candidate answers for each question. This answer pool was produced by including the correct answer and randomly selected candidates from the complete set of unique answers. The top-1 accuracy of the answer selection is reported.

\textbf{Yahoo QA} is an open-domain community-based dataset, which is collected from Yahoo Answers and pre-processed by \cite{DBLP:conf/sigir/TayPLH17}. The lengths of sentences are from 5 to 50. There are 5 candidate answers under each question, including 4 generated negative samples and 1 ground truth. The same metrics as \cite{DBLP:conf/sigir/TayPLH17,DBLP:conf/wsdm/TayTH18} are adopted for evaluation, including Precision@1 (denoted as P@1) and MRR.

\subsubsection{Implementation Details}\label{subsubsection512}
Pre-trained GloVE embeddings\footnote{http://nlp.stanford.edu/data/glove.6B.zip} of 300 dimensions are adopted as word embeddings. We use a subset of Freebase (FB5M\footnote{https://research.facebook.com/researchers/1543934539189348}) as our KG, which includes 4,904,397 entities, 7,523 relations, and 22,441,880 facts. TagMe\footnote{https://github.com/marcocor/tagme-python
} is applied to derive the entity mentions with the confidence score above the threshold of 0.2 for each sentence. Since there is no ground truth for the entity mention extraction, the confidence score is chosen by a small validation set of 100 sentences with human evaluation based on F1 score. Then the entity mention is matched to the certain entity candidates in the Freebase to construct the knowledge sequence. TransE~\cite{DBLP:conf/nips/BordesUGWY13} is adopted as the knowledge embedding method to generate the knowledge embeddings for all the models. OpenKE\footnote{https://github.com/thunlp/OpenKE} is employed to implement TransE with the default settings. We extract top-K entity candidates for each entity mention in the experiment, where $K=5$ is tuned on the validation set.

For all of our proposed models, we apply the same parameter settings. The LSTM hidden layer size and the final hidden layer size are both set to 200. The learning rate and the dropout rate are set to 0.0005 and 0.5 respectively. We train our models in batches with the size of 64. All other parameters are randomly initialized from [-0.1, 0.1]. The model parameters are regularized with a L2 regularization strength of 0.0001. The maximum length of sentence is set to be 40. In the knowledge module, the
width of the convolution filters is set to be 2 and 3, and the number of
convolutional feature maps and the attention sizes are set to be 200. The overall parameter quantity in KNN, KANN and CKANN is about 5.69M, 14.88M, and 20.18M, respectively.

\subsubsection{Baseline Methods}
We make a very extensive comparison with the state-of-the-arts methods. Since some IR-based methods and machine learning methods have been proven to be less effective on answer selection~\cite{DBLP:conf/kdd/WangJY17}, we only adopt methods based on deep learning model to conduct comparisons as follows:

\begin{itemize}
    \item \textbf{CNN}. \citet{DBLP:conf/sigir/SeverynM15} propose a siamese CNN architecture for ranking   QA pairs with overlap features and similarity scores as additional features.
    \item \textbf{CNN-Cnt}. ~\citet{DBLP:conf/emnlp/YangYM15} combine CNN model with the two word matching features by training a logistic regression classifier.
    \item \textbf{Bi-LSTM+BM25}. \citet{DBLP:conf/acl/WangN15} employ a stacked Bi-LSTM network to sequentially encode words from question and answer sentences and incorporates BM25 score as an additional feature.
    \item \textbf{Attentive LSTM}. \citet{DBLP:conf/acl/TanSXZ16} propose an Attentive LSTM model with a simple but effective attention mechanism for the purpose of improving the semantic representations for the answers based on the questions
    \item \textbf{ABCNN}. \citet{DBLP:journals/tacl/YinSXZ16} integrate attention schemes into CNN models, which takes mutual influence between sentences into consideration.
    \item \textbf{IARNN}. \citet{DBLP:conf/acl/WangL016} present novel RNN models that integrate attention scheme before RNN hidden representation. \textbf{IARNN-Gate} adds attention information to some active gates to influence the hidden representation. \textbf{IARNN-Occam} applies Occam’s Razor into RNN model to incorporate attention regulation. 
    \item \textbf{AP-LSTM}. \citet{DBLP:journals/corr/SantosTXZ16} propose a novel attention-based LSTM model with attentive pooling, a two-way attention mechanism that information from the question and the answer influences the representation of each other mutually. 
    \item \textbf{NCE}. \citet{DBLP:conf/cikm/RaoHL16} propose a pairwise ranking approach that employs Noise-Contrastive Estimation (NCE) for training model to discriminate a good sample from its neighboring bad samples.
    \item \textbf{HD-LSTM}. \citet{DBLP:conf/sigir/TayPLH17} extensd the Siamese LSTM network with holographic composition to model the relationship between question and answer representations.
    \item \textbf{RNN-POA}. \citet{DBLP:conf/sigir/ChenHHHA17} propose a RNN model with position-aware attention mechanism, which incorporates the positional context of the question words into the answers’ attentive representations.
    \item \textbf{Conv-RNN}. \citet{DBLP:conf/kdd/WangJY17} propose a hybrid framework of attention-based Convolutional Recurrent Neural Network (Conv-RNN) with a similar attention mechanism as \cite{DBLP:conf/acl/TanSXZ16} but in the input of the RNN layer.
    \item \textbf{CTRN}. \citet{DBLP:conf/aaai/TayTH18} learn temporal gates for QA pairs, jointly influencing the learned representations in a pairwise manner.
    \item \textbf{CNN$_\mathbf{WO+SO}$}. \citet{DBLP:conf/emnlp/NicosiaM18} model semantic phenomena in text into CNN model~\cite{DBLP:conf/sigir/SeverynM15} to encode matching features directly in the word representations.
    \item \textbf{HyperQA}. \citet{DBLP:conf/wsdm/TayTH18} propose a pairwise ranking objective that models the relationship between question and answer embeddings in Hyperbolic space instead of Euclidean space.
    \item \textbf{HCAN}. \citet{DBLP:conf/emnlp/RaoLTYSL19} propose a hybrid co-attention network that bridges the gap between relevance matching and semantic matching.
    \item \textbf{Hard Negatives (HN)}. \citet{DBLP:conf/emnlp/KumarGMR19} couples a siamese network with hard negative, including two variants: \textbf{LSTM-CNN + HN} and \textbf{TFM + HN}.
    \item \textbf{SD}. \citet{DBLP:conf/sigir/0013PLXZX20} propose a similarity aggregation method to rerank the results produced by different baseline neural networks, including three variants: \textbf{SD(CNN+TFM)}, \textbf{SD(CNN+BiLSTM)}, and \textbf{SD(BiLSTM+TFM)}.
\end{itemize}

\subsection{Answer Selection Results}\label{subsection52}
\subsubsection{Results on Factoid QA}

\begin{table*}
\fontsize{10}{11}\selectfont
\centering
  \caption{Experimental Results on Factoid QA. $^*$ indicates that  CKANN is better than KANN with statistical significance (measured by significance test at $p<0.05$).}
  \begin{tabular}{lcccccc}
    \toprule
	\multirow{2}{*}{Model} & \multicolumn{2}{c}{TREC QA(original)} & \multicolumn{2}{c}{TREC QA(cleaned)} & \multicolumn{2}{c}{WikiQA} \\
	\cmidrule(lr){2-3} \cmidrule(lr){4-5} \cmidrule(lr){6-7} 
	 & MAP & MRR & MAP & MRR & MAP & MRR  \\
    \midrule
    \textbf{CNN} (2015)~\cite{DBLP:conf/sigir/SeverynM15} & 0.746 & 0.808&-&-&-&-\\
    \textbf{CNN-Cnt} (2015)~\cite{DBLP:conf/emnlp/YangYM15} &-&-&0.695&0.763& 0.652 & 0.665 \\
    \textbf{Bi-LSTM+BM25} (2015)~\cite{DBLP:conf/acl/WangN15}&-&- & 0.713 & 0.791 &-&-\\
    \textbf{Attentive LSTM} (2016)~\cite{DBLP:conf/acl/TanSXZ16}&0.753& 0.830&-&-&-&-\\
    \textbf{ABCNN} (2016)~\cite{DBLP:journals/tacl/YinSXZ16}&-&-&-&-& 0.692& 0.711\\
    \textbf{IARNN-Gate} (2016)~\cite{DBLP:conf/acl/WangL016} & 0.737 & 0.821&-&-& 0.726& 0.739\\
    \textbf{IARNN-Occam} (2016)~\cite{DBLP:conf/acl/WangL016} & 0.727& 0.819&-&-& \underline{0.734}&\underline{0.742}\\
    \textbf{AP-LSTM} (2016)~\cite{DBLP:journals/corr/SantosTXZ16}&-&-&0.753&0.851 & 0.689 & 0.696\\
    \textbf{NCE-WordLevel} (2016)~\cite{DBLP:conf/cikm/RaoHL16} & 0.764& 0.827&0.762& 0.854 & 0.693& 0.710\\
    \textbf{NCE-SentLevel} (2016)~\cite{DBLP:conf/cikm/RaoHL16} & \underline{0.780} & 0.834&\underline{0.801} &0.877 & 0.701 & 0.718\\
	\textbf{HD-LSTM} (2017)~\cite{DBLP:conf/sigir/TayPLH17} & 0.750 & 0.815 &-&-&-&-\\
	\textbf{RNN-POA} (2017)~\cite{DBLP:conf/sigir/ChenHHHA17}&-&-&0.781&0.851 &0.721& 0.731\\
	\textbf{CTRN} (2018)~\cite{DBLP:conf/aaai/TayTH18} &0.771& 0.838&-&-&-&-\\
    \textbf{CNN$_\mathbf{WO+SO}$} (2018)~\cite{DBLP:conf/emnlp/NicosiaM18}&-&-&0.779& 0.849&0.722 &0.739 \\
    \textbf{HyperQA} (2018)~\cite{DBLP:conf/wsdm/TayTH18}& 0.770& 0.825& 0.784& 0.865&0.712& 0.727\\
    \textbf{HCAN} (2019)~\cite{DBLP:conf/emnlp/RaoLTYSL19}&0.774 &\underline{0.843}&-&-&-&- \\
    \textbf{SD(CNN+TFM)} (2020)~\cite{DBLP:conf/sigir/0013PLXZX20} &-&-&0.783&\underline{0.878}&0.691&0.703\\ \textbf{SD(CNN+BiLSTM)} (2020)~\cite{DBLP:conf/sigir/0013PLXZX20}&-&-&0.780&\underline{0.878}&0.700&0.710\\
    \textbf{SD(BiLSTM+TFM)} (2020)~\cite{DBLP:conf/sigir/0013PLXZX20}&-&-&0.755&0.841&0.704&0.712\\
    \midrule
   KNN & 0.781 & 0.831&0.782&0.865& 0.721 & 0.736   \\
   KANN w/ self-attention & 0.790 & 0.841&0.793&0.877& 0.731 & 0.742   \\
   - w/ co-attention &0.792&0.844&0.804&0.885 &0.734 &0.752\\
   - w/ multi-view attention &0.802&0.852&0.812&0.890 &0.735 &0.754\\
   \midrule
		CKANN-2 & 0.812 & 0.856 & 0.824 & 0.889 & \textbf{0.736} & 0.746  \\
		CKANN-3 & 0.796 & 0.841 & 0.817 & 0.882 & 0.729 & 0.733  \\
		CKANN-L & 0.808 & 0.853 & 0.819 & 0.882 & 0.728 & 0.739  \\
		\midrule
		CKANN & \textbf{0.819}$^*$ & \textbf{0.863}$^*$ & \textbf{0.825}$^*$ & \textbf{0.891}$^*$ & 0.732$^*$ & \textbf{0.755}$^*$  \\
  \bottomrule
\label{exp}
\end{tabular}
\end{table*}

\begin{table*}
\centering
  \caption{Experimental Results on Non-Factoid QA. $^*$ indicates that  CKANN is better than KANN with statistical significance (measured by significance test at $p<0.05$).}
  \begin{tabular}{lccccc}
    \toprule
	\multirow{2}{*}{Model} & \multicolumn{2}{c}{Yahoo QA}& \multicolumn{3}{c}{InsuranceQA} \\
	\cmidrule(lr){2-3} \cmidrule(lr){4-6} 
	  & P@1 & MRR & DEV & TEST1 & TEST2 \\
    \midrule
    \textbf{Attentive LSTM} (2016)~\cite{DBLP:conf/acl/TanSXZ16}&-&-&0.689& 0.690 &0.648\\
    \textbf{IARNN-Gate} (2016)~\cite{DBLP:conf/acl/WangL016} & -&-&0.700& 0.701& 0.628\\
    \textbf{IARNN-Occam} (2016)~\cite{DBLP:conf/acl/WangL016} &-&-&0.691& 0.689& 0.651\\
    \textbf{AP-LSTM} (2016)~\cite{DBLP:journals/corr/SantosTXZ16}&-&-&0.684 &0.717 &0.664\\
	\textbf{Conv-RNN} (2017)~\cite{DBLP:conf/kdd/WangJY17} &-&- &0.717 &0.714 &0.683\\
	\textbf{HD-LSTM} (2017)~\cite{DBLP:conf/sigir/TayPLH17} &0.557&0.735&-&-&- \\
	\textbf{CTRN} (2018)~\cite{DBLP:conf/aaai/TayTH18} &0.601&0.755&-&-&- \\
	\textbf{HyperQA} (2018)~\cite{DBLP:conf/wsdm/TayTH18} &\underline{0.683}&\underline{0.801}&-&-&- \\
    \textbf{LSTM-CNN + HN} (2019)~\cite{DBLP:conf/emnlp/KumarGMR19}&-&-&0.725&0.733&0.691\\
    \textbf{TFM + HN} (2019)~\cite{DBLP:conf/emnlp/KumarGMR19}&-&-&\underline{0.757}&\underline{0.756}&\underline{0.734}\\
    \midrule
   KNN &  0.725 & 0.844 &0.713&0.715&0.688  \\
   KANN w/ self-attention & 0.745 & 0.860 &0.717&0.719&0.694  \\
   - w/ co-attention &0.766 &0.871&0.724&0.728&0.701 \\
   - w/ multi-view attention &0.789 &0.884&0.731&0.734&0.712 \\
   \midrule
		CKANN-2 & 0.835&0.901&0.751&0.752&0.744 \\
		CKANN-3 & 0.836 & 0.901&\textbf{0.764}&0.751&0.744 \\
		CKANN-L & 0.842 & \textbf{0.906}&0.758&0.759&0.749 \\
		\midrule
		CKANN  &\textbf{0.844}$^*$&0.902$^*$&0.761$^*$&\textbf{0.763}$^*$&\textbf{0.751}$^*$ \\
  \bottomrule
\label{exp-non}
\end{tabular}
\end{table*}

The experimental results on the factoid QA datasets, namely TREC QA and WikiQA , are summarized in Table~\ref{exp}. We observe that KANN substantially and consistently outperforms the existing methods by a noticeable margin on 3 out of 4 datasets. For instance, on the original TREC QA dataset, KANN improves 2\% on MAP over these baselines. Notice that even the KNN without knowledge-aware attention mechanism can achieve competitive results with those strong baseline methods, which demonstrates that it is effective to incorporate external knowledge from knowledge graph to make a better performance on answer selection task. Besides, for context-based learning module, even though we just employ the basic Bi-LSTM model, instead of other deep learning models with complicated architecture, our proposed method achieves a better performance than these context-based methods. This result indicates that background knowledge from a well-constructed knowledge graph can directly capture deep semantic information without a complex semantic analysis model. As for \textbf{RQ1}, \textit{answer selection in factoid QA task actually can benefit from incorporating external knowledge from knowledge graph.} For the proposed knowledge-aware attention mechanism, the experimental results show that knowledge-aware co-attention mechanism outperforms self-attention by averagely 1\% on all the datasets. And multi-view attention further improves the performance by considering the interactions from various perspectives. This provides an answer to \textbf{RQ2}: \textit{(i) the interaction between the textual information and the knowledge information can improve the model performance; (ii) it is better to take into account the context-knowledge interaction from both question and answer (Knowledge-aware Co-Attention), rather than consider separately (Knowledge-aware Self-Attention); (iii) multi-view attention scheme provides the most effective way to integrate the interactions between context and knowledge information from various perspectives.}

For CKANN, we implement three basic models by using the similar settings as CKANN but only one of three types of Entity Graphs ($p=2,3,L_k$) respectively, namely CKANN-2, CKANN-3, and CKANN-L. We can observe that our proposed model substantially and consistently outperforms the existing methods and KANN by a noticeable margin. For instance, our model improves 2.2\% MAP and 1.3\% MRR on TREC QA dataset. CKANN-2, CKANN-3 and CKANN-L achieve competitive results, but fluctuate slightly on different datasets. This is because the $p$ value is related to the connection between the entity nodes in the Entity Graph. When the $p$ value is too low, the connections between the entity nodes are not closely linked. When the $p$ value is too high, semantic noise may be generated due to the connection of extraneous entities that are far apart. By applying mean-pooling operation to summarize and smooth three types of Entity Graphs with different $p$, our model CKANN achieves stable performance and shows robustness and effectiveness on different datasets. KNN and KANN directly apply the fixed knowledge embeddings, which are not closely related to the entire question/answer sentences. To address this issue, our model learns contextualized knowledge representation to enrich the semantic relationship among knowledge and therefore obtains a better result. This result shows an evidence for \textbf{RQ3}: \textit{applying GCN to learn the contextualized knowledge representation is more helpful than explicitly adopting the fixed pre-trained knowledge embeddings for the knowledge-based learning module.}

\subsubsection{Results on Non-factoid QA}
The experimental results on two non-factoid QA datasets, namely Yahoo QA and InsuranceQA, are presented in Table~\ref{exp-non}. We observe that similar to the results on factoid QA datasets, the proposed knowledge-aware neural networks substantially and consistently outperform state-of-the-art context-based models by a large margin.  As for \textbf{RQ1}, \textit{answer selection in non-factoid QA task can also take advantages of incorporating external knowledge from knowledge graph.} Compared with the results on factoid QA datasets, we notice that there is a more distinguishable performance gap between different knowledge-aware attention mechanism and different knowledge-aware neural frameworks. The basic KNN model can achieve competitive results as these strong baseline models, indicating that the knowledge information is also of great importance on non-factoid answer selection tasks. As for factoid QA, all the candidate answers are likely to talk about the same topic or even the same range of knowledge, while on the non-factoid QA, different candidate answers are more differentiated in the knowledge-based representational learning. Especially, CKANN, which takes into account a subgraph of mentioned and related knowledge from the sentence, achieves superior performance on the non-factoid QA datasets. These results answer \textbf{RQ1} from a different angle: \textit{The difference degree of mentioned knowledge aids in selecting the most related answers for the given question, which is underutilized in those context-based methods.}

\begin{table*}
\fontsize{8}{9.5}\selectfont
\centering
\caption{\label{table:bert} Experimental results on Comparison with BERT-based Models. $^*$ indicates that the model is better than BERT$_{\text{base}}$ model with statistical significance (measured by significance test at $p<0.05$).}
\begin{tabular}{lcccccc}
    \toprule
    \multirow{2}{*}{Model}&\multicolumn{2}{c}{TREC QA(cleaned)}& \multicolumn{2}{c}{WikiQA}& \multicolumn{2}{c}{Yahoo QA}\\ 
	\cmidrule(lr){2-3}\cmidrule(lr){4-5}\cmidrule(lr){6-7}
	&MAP&MRR&MAP&MRR&P@1&MRR\\
	\midrule
	BERTSel (2019)~\cite{DBLP:journals/corr/abs-1905-07588}& 0.877& 0.927& 0.753& 0.770& - &\underline{0.942}\\
    \midrule
    BERT$_{\text{base}}$ (2019)~\cite{DBLP:conf/emnlp/LaiTBK19} &0.877& 0.922& 0.810& 0.827&-&-\\
    -TRF (2019)~\cite{DBLP:conf/emnlp/LaiTBK19}& 0.886 ($\uparrow$.009)& 0.926 ($\uparrow$.004)& 0.813 ($\uparrow$.003)& 0.828 ($\uparrow$.001)&-&-\\
    -GSAMN (2019)~\cite{DBLP:conf/emnlp/LaiTBK19}& \underline{0.906} ($\uparrow$.029)& \underline{0.949} ($\uparrow$\textbf{.027})& \underline{\textbf{0.821}} ($\uparrow$.011)& \underline{\textbf{0.832}} ($\uparrow$.005)&-&-\\
    \midrule
    BERT$_{\text{base}}$ (2020)~\cite{DBLP:conf/aaai/XuL20}&-&-& 0.807&  0.818& 0.748&  0.827 \\
    HAS (2020)~\cite{DBLP:conf/aaai/XuL20}&-&-&0.810 ($\uparrow$.003)& 0.822 ($\uparrow$.004)&0.739 ($\downarrow$.009)& 0.821 ($\downarrow$.006)\\
    \midrule
    BERT$_{\text{base}}$& 0.872& 0.924&  0.772& 0.783 &0.842&0.940\\
    KNN (BERT$_{\text{base}}$)& 0.900$^*$($\uparrow$.028)& 0.938$^*$($\uparrow$.014)&0.784$^*$($\uparrow$.012) &0.796$^*$($\uparrow$.013) &0.846 ($\uparrow$.004)&0.939 ($\downarrow$.001)\\
    KANN (BERT$_{\text{base}}$)& 0.905$^*$($\uparrow$.033)& 0.946$^*$($\uparrow$.022)& 0.792$^*$($\uparrow$.020)& 0.806$^*$($\uparrow$.023)&\textbf{0.852}$^*$($\uparrow$\textbf{.010})&0.947$^*$($\uparrow$.007)\\
    CKANN (BERT$_{\text{base}}$) & \textbf{0.909}$^*$($\uparrow$\textbf{.037})& \textbf{0.952}$^*$($\uparrow$.026)& 0.802$^*$($\uparrow$\textbf{.030})& 0.819$^*$($\uparrow$\textbf{.036})& 0.850$^*$($\uparrow$.008)& \textbf{0.948}$^*$($\uparrow$\textbf{.008})\\
    \bottomrule
\end{tabular}

\end{table*}

\subsubsection{Comparison with BERT-based Models}
In order to investigate whether the external knowledge is still helpful when adopting pre-trained language models, we replace the GloVe embeddings with a pre-trained BERT sentence encoder for the proposed models. We adopt several state-of-the-art BERT-based answer selection methods for comparison:
\begin{itemize}
    \item \textbf{BERTSel}. \citet{DBLP:journals/corr/abs-1905-07588} explore the performance of fine-tuning BERT for answer selection. 
    \item \textbf{BERT-TRF}. \citet{DBLP:conf/emnlp/LaiTBK19} use two Transformer layers on top of BERT.
    \item \textbf{BERT-GSAMN}. \citet{DBLP:conf/emnlp/LaiTBK19} propose a gated self-attention memory network with pre-trained BERT sentence encoder.
    \item \textbf{HAS}. \citet{DBLP:conf/aaai/XuL20} propose a hashing strategy to reduce the memory cost for storing the matrix representations of answers when adopting complex encoders like BERT.
\end{itemize}

For a fair comparison, we report the performance of these models using the original BERT$_\text{base}$ as the pre-trained language model in Table~\ref{table:bert}. 
Compared with these state-of-the-art BERT-based methods, we observe that the proposed models with fine-tuned BERT encoder can still generally achieve better performance on these datasets, and achieve a new SOTA result on TREC QA and Yahoo QA. Previous studies on fine-tuning BERT~\cite{DBLP:journals/corr/abs-2002-06305,DBLP:journals/corr/abs-2006-05987} show that two sources of randomness, including the weight initialization of the new output layer and the data order in the stochastic fine-tuning optimization, can influence the fine-tuning results significantly, especially on small datasets (i.e., < 10K examples). Therefore, it can be observed that even the results of vanilla fine-tuned BERT on WikiQA are reported differently to a great extent~\cite{DBLP:journals/corr/abs-1905-07588, DBLP:conf/emnlp/LaiTBK19, DBLP:conf/aaai/XuL20} due to the small amount of samples on WikiQA datasets ($\approx$ 10K examples). More importantly, the results show that the BERT-based model can also be benefited from incorporating external knowledge from a knowledge graph with our proposed knowledge module to a great extent. Compared with other design, our proposed context-knowledge interaction learning framework achieves the most significant improvement over vanilla fine-tuned BERT model.

\subsection{Ablation Study}\label{subsection56}
\subsubsection{Effect of Additional Features}
As mentioned in Section 3.3, there are several additional features incorporated into the overall framework, including bilinear similarity score $sim$ and the overlap features $X_{feat}$. Since we follow some previous works~\cite{DBLP:conf/sigir/SeverynM15,DBLP:conf/sigir/TayPLH17} to employ these additional features, we hope to investigate that these additional features can contribute to a better performance or dominate the whole result.  

From Table~\ref{feature} we can observe that: (1) Only using overlap features can achieve somewhat reasonable results, but there is still a big gap, about 10\%, between the results and deep learning models. However, the ablation tests in terms of discarding overlap features show that it makes an improvement about 1-3\% on the final results if these overlap features are incorporated. These results show that the overlap features are effective to boost the performance but not dominating the overall results. (2) Compared with overlap features, incorporating similarity scores has a trivial effect on the overall performance, approximately 1\%.  In addition, different kinds of similarity scores achieve similar results, but among these three kinds of similarity scores, \emph{dot} and \emph{bilinear} achieve the best performance.

\begin{table}
\centering
  \caption{Effect of Additional Features}
  \begin{tabular}{ccccccc}
    \toprule
   \multirow{2}{*}{$s_q,s_a$} & \multirow{2}{*}{$X_{feat}$} & \multirow{2}{*}{$sim$} &  \multicolumn{2}{c}{TREC QA(original)}&\multicolumn{2}{c}{WikiQA}\\
    \cmidrule(lr){4-5} \cmidrule(lr){6-7} 
    & &&MAP& MRR&MAP & MRR\\
   \midrule
   -&+&-&0.737&0.806&0.628&	0.638\\
    \midrule
   \multirow{4}{*}{+}&\multirow{4}{*}{-}&-&0.766&0.817&0.717&0.733\\
   &&bilinear&0.760&0.821&0.714&0.732\\
   &&cosine&0.762&0.810&0.715&0.735\\
   &&dot&\textbf{0.769}&\textbf{0.826}&\textbf{0.724}&\textbf{0.740}\\
    \midrule
    \multirow{4}{*}{+}&\multirow{4}{*}{+}&-&0.783&0.835&0.726&0.742\\
    &&bilinear&0.792&\textbf{0.844}&\textbf{0.734} &\textbf{0.752}\\
   &&cosine&0.780&0.841&0.726&0.743\\
   &&dot&\textbf{0.795}&0.839&0.732&0.751\\
  \bottomrule
\end{tabular}
\label{feature}
\end{table}

\subsubsection{Ablation Study on CKANN}

\begin{table}
	\centering
	\caption{\label{table:ablation} Ablation study}
	\begin{tabular}{lcccccccccc}
		\toprule
		\multirow{2}{*}{Model}&\multicolumn{2}{c}{TREC QA(original)}&\multicolumn{2}{c}{Wiki QA}&\multicolumn{2}{c}{Yahoo QA}&\multicolumn{3}{c}{InsuranceQA} \\ 
		\cmidrule(lr){2-3}\cmidrule(lr){4-5}\cmidrule(lr){6-7}\cmidrule(lr){8-10}
		&MAP&MRR&MAP&MRR&P@1&MRR&DEV&TEST1&TEST2\\
		\midrule
		CKANN & \textbf{0.819} & \textbf{0.863} & \textbf{0.732} & \textbf{0.755}&\textbf{0.844}&\textbf{0.902}&\textbf{0.761}&\textbf{0.763}&\textbf{0.751}  \\
		\midrule
		w/o Know-Att & 0.805 & 0.849 & 0.714 & 0.728 & 0.835 & 0.902  & 0.759 & 0.747 & 0.751\\
		w/o GCN & 0.797 & 0.845 & 0.715 & 0.731 & 0.837 & 0.901 & 0.751 & 0.759 & 0.741\\
		w/o KGs & 0.782 & 0.825  & 0.706 & 0.723 & 0.834 & 0.900 & 0.753 & 0.753 & 0.743\\
		\bottomrule 
	\end{tabular}
\end{table}

For a thorough comparison, we report the ablation test to analyze the improvements contributed by each part of our model: (i) \textbf{w/o Know-Att}: We remove the multi-view attention mechanism in our model. (ii) \textbf{w/o KGs}: We discard the background information from KGs. (iii) \textbf{w/o GCN}: We replace the contextualized knowledge representation learned by GCN with fixed pre-trained TransE knowledge embeddings.

The results are shown in Table~\ref{table:ablation}. 
We can see that applying multi-view attention mechanism can contribute to model performance, since attention mechanism can perceive the importance of different words and knowledge. Furthermore, the proposed novel multi-view attention can capture not only the semantic relations between words of QA pairs, but also the interrelation between background knowledge of sentences.

Compared with ``CKANN'' and ``w/o KGs'', we can conclude that by introducing KG's background information and learning knowledge-aware sentence representations, our model has achieved better performance (improving 3.7\% MAP and 3.8\% MRR on TREC QA). 
And our model outperforms "w/o GCN" by a noticeable margin, demonstrating the contribution of learning contextualized knowledge representation in capturing the semantic relationship of the entire sentence. In addition, we observe that the knowledge-enhanced representation learning makes more contribution to the performance boosting in factoid QA than that in non-factoid QA.

\subsection{Applicability Analysis}\label{subsection53}
As mentioned in Sections 3 \& 4, we aim to develop a general neural network framework for incorporating external knowledge into answer selection model. In our proposed framework, Bi-LSTM is adopted as the context-based learning module for its simplicity and efficiency. In this section, we implement the overall neural network framework with several LSTM-based deep learning models, besides Bi-LSTM, to demonstrate the applicability and universality of the proposed method. We implement all these models into the proposed framework as the context module. In order to analyze the effectiveness of different factors, we also report the basic neural frameworks in terms of discarding knowledge-aware attention mechanism (KNN) and knowledge graph information (denoted as None), respectively. The context-based neural models we investigate include:

\begin{itemize}
    \item \textbf{Att-LSTM}, an Attentive LSTM model~\cite{DBLP:conf/acl/TanSXZ16} with a simple but effective attention mechanism for the purpose of improving the semantic representations for the answers based on the questions; 
    \item \textbf{AP-LSTM}, an AP-LSTM model~\cite{DBLP:journals/corr/SantosTXZ16} with attentive pooling, a two-way attention mechanism that information from the question and the answer can directly influence the computation of each other’s representations ; 
    \item \textbf{Conv-RNN}, a hybrid framework of attention-based Convolutional Recurrent Neural Network (Conv-RNN)~\cite{DBLP:conf/kdd/WangJY17} with a similar attention mechanism as Att-LSTM but in the input of the RNN. Note that, unlike the original paper of Conv-RNN, we employ Bi-LSTM as the RNN model instead of the Bi-GRU.
\end{itemize}

For the details of these models, please refer to the original papers. The experimental results on TREC QA and WikiQA are summarized in Table~\ref{app}.

\begin{table}
\centering
  \caption{Applicability of the proposed frameworks}
  \begin{tabular}{cccccc}
    \toprule
   Context & \multirow{2}{*}{Method} & \multicolumn{2}{c}{TREC QA(original)}&\multicolumn{2}{c}{WikiQA}\\
    \cmidrule(lr){3-4} \cmidrule(lr){5-6} 
    Module & &MAP& MRR&MAP & MRR\\
     \midrule
   \multirow{4}{*}{Att-BiLSTM}&None&0.735&0.792&0.689&0.698\\
   &KNN&0.759&0.801&0.701&0.711\\
  &KANN&0.770&0.822&0.712&0.720\\
  &CKANN&0.818&0.854&0.722&0.749\\
   \midrule
  \multirow{4}{*}{AP-BLSTM}&None&0.760&0.810&0.709&0.726\\
   &KNN&0.774&0.835&0.714&0.734\\
   &KANN&0.783&0.842&0.730&0.746\\
   &CKANN&0.823&0.868&0.730&0.753\\
    \midrule
   \multirow{4}{*}{Conv-RNN}&None&	0.785&0.824&0.710&0.724\\
   &KNN&0.789&0.837&0.721&0.735\\
   &KANN&0.793&0.841&0.735&0.750\\
   &CKANN&0.822&0.870&0.737&0.759\\
  \bottomrule
\end{tabular}
\label{app}
\end{table}

Generally, we obtain similar results as Bi-LSTM with other models. For all the models, it makes a significant performance boost to incorporate external knowledge into the overall architecture, and knowledge-aware attention mechanism further improves the final results, which demonstrates that the proposed framework works on all the given models. 

In specific, we observe that scaling up the neural network can easily improve the performance without external knowledge. Compared with the basic Bi-LSTM model in Table~\ref{exp}, AP-BLSTM slightly improves the performance of Bi-LSTM, while Att-BLSTM obtains a similar performance. Conv-RNN achieves the best result with the basic setting. Even the basic Bi-LSTM model with external knowledge integrated by the proposed knowledge module achieves competitive results with other complex deep learning models, which demonstrates the effectiveness of incorporating background knowledge into ranking QA pairs. This is within our expectation since KG introduces background knowledge beyond the context to enrich overall sentence representations, while the knowledge-aware attention mechanism further enhances mutual representational learning of QA sentences. Moreover, CKANN performs much better than KNN framework with different kinds of context-based learning module, by adopting GCN to learn the contextualized knowledge representations, which also answers \textbf{RQ3}.

For KNN, KANN and CKANN, the experimental results show that there is little difference between different models in context-based learning module, which indicates that the knowledge module makes more contribution to the overall performance and also demonstrate the effectiveness of incorporating external knowledge into answer selection model. This result provides the answer to \textbf{RQ4}: \textit{the proposed framework can be easily and effectively adapted to diverse existing context-based answer selection models, and achieve superior performance by incorporating external knowledge from knowledge graph.}

\subsection{Analysis of External Knowledge Integration}\label{subsection54}

\subsubsection{Comparison of Knowledge Embedding Methods}
To compare different kinds of knowledge embedding methods and demonstrate the superiority of the proposed contextualized knowledge representation learning method, we conduct the experiments on knowledge embedding methods. Based on the proposed knowledge-aware neural network framework, KNN and KANN, we first input with zero vectors as knowledge embeddings as the baseline, namely \textbf{None}. Then we randomly generate knowledge embeddings with 64 dimensions to observe the effect of incorporating knowledge into answer selection models, namely \textbf{Random}. 
Besides TransE~\cite{DBLP:conf/nips/BordesUGWY13}, we adopt several popular network or knowledge embedding methods for comparison, including Deepwalk~\cite{DBLP:conf/kdd/PerozziAS14}, LINE~\cite{DBLP:conf/www/TangQWZYM15},   TransD~\cite{DBLP:conf/acl/JiHXL015}, TransH~\cite{DBLP:conf/aaai/WangZFC14},  HolE~\cite{DBLP:conf/aaai/NickelRP16} and ComplEx~\cite{DBLP:conf/icml/TrouillonWRGB16}.

\begin{itemize}
    \item \textbf{Deepwalk} exploits local information obtained from truncated random walks to learn latent representations of vertices in a network by treating walks as the equivalent of sentences.

    \item \textbf{LINE} optimizes a novel objective function that involves both the local and global network structures, and employs an edge-sampling algorithm to address the limitation of the stochastic gradient descent and improves both the effectiveness and the efficiency of the inference. 

    \item \textbf{TransE} models relations by interpreting them as translations operating on the low-dimensional embeddings of the entities.

    \item \textbf{TransD} simplifies TransR~\cite{DBLP:conf/aaai/LinLSLZ15} by decomposing the projection matrix into a product of two vectors. TransR builds entity and relation embeddings in separate entity space and relation spaces, which is different from TransE.

    \item \textbf{TransH} makes a good trade-off between model efficiency and capacity to preserve different kinds of mapping properties of relations,  by modeling a relation as a hyperplane together with a translation operation on it.

    \item \textbf{HolE} employs correlation as the compositional operator to capture interactions and learns compositional vector space representations of entire knowledge graphs.

    \item \textbf{ComplEx} only uses the Hermitian dot product, the complex counterpart of the standard dot product between real vectors, and introduces complex-valued embeddings to better model asymmetric relations.
\end{itemize}

For implementation, we employ OpenNE~\footnote{https://github.com/thunlp/OpenNE} to generate Deepwalk and LINE embeddings with the default settings. OpenKE is also employed to implement TransD, TransH, HolE and ComplEx with the default settings.

Table~\ref{kemethod} collects the experiment results on knowledge embedding (KE) methods with two proposed neural network frameworks, KNN and KANN, and compare them with the proposed contextualized knowledge representation learning method. There are multiple interesting observations: 

\begin{table}
\centering
  \caption{Comparison of Knowledge Embedding Methods}
  \begin{tabular}{cccccc}
    \toprule
   \multirow{2}{*}{Model} & \multirow{2}{*}{KE Method} & \multicolumn{2}{c}{TREC QA(original)}&\multicolumn{2}{c}{WikiQA}\\
    \cmidrule(lr){3-4} \cmidrule(lr){5-6} 
    & &MAP& MRR&MAP & MRR\\
   \midrule
    &None&0.750&0.804&0.688&0.700\\
   &Random&0.769&0.825&0.711&0.726\\
    &Deepwalk&0.774&0.826&0.713&0.731\\
    &LINE&0.772&0.828&0.711&0.731\\
    KNN&TransE &\textbf{ 0.784} & 0.833& 0.717 & 0.736\\
    (Bi-LSTM)&TransD& 0.782 & 0.834& 0.716 & 0.734\\
    &TransH&0.783 &\textbf{ 0.837}& \textbf{0.721} & \textbf{0.736}\\
    &HolE& 0.781 & 0.832& 0.718 & 0.733\\
    &ComplEx& 0.780 & 0.831& 0.717 & 0.732\\
    \midrule
   &None&0.760&0.810&0.709&0.726\\
   &Random&0.768&0.830&0.710&0.729\\
     &Deepwalk&0.785&0.829&0.725&0.745\\
     &LINE&0.784&0.831&0.724&0.741\\
     KANN&TransE&\textbf{0.792}&\textbf{0.844}&0.734 &\textbf{0.752}\\
     (Bi-LSTM)&TransD&0.787&0.843&0.730&0.746\\
    &TransH&0.790&0.843&\textbf{0.735}&0.749\\
    &HolE&0.789&0.843&0.731&0.747\\
    &ComplEx&0.790&0.842&0.730&0.747\\
    \midrule
    CKANN&Random&0.782 & 0.825  & 0.706 & 0.723\\
   (Bi-LSTM)&TransE&\textbf{0.819}&\textbf{0.863}&\textbf{0.732}&\textbf{0.755}\\
  \bottomrule
\end{tabular}
\label{kemethod}
\end{table}

(1) KANN outperforms KNN by 1-3\% with all kinds of KE methods in two datasets, which also validates the effectiveness of the knowledge-aware mechanism as well as testify the applicability of the overall framework.

(2) Initializing the entity with random generated embeddings outperforms that with zero vectors by about 1-2\% in both two frameworks and two datasets. With zero vectors as the knowledge embeddings, the knowledge module is disabled, while with random embeddings, the knowledge module simply labels the entity in sentences without any extra information. This result indicates that the knowledge (entities) are important in answer selection task. Being aware of the knowledge can make a better performance. 

(3) Knowledge embedding methods further improve the performance by 2-3\% over random generated embeddings. This result shows than knowledge embedding methods effectively provides more informative representations for entities, since these methods incorporate the structure information of a knowledge graph as a graph or a network into the knowledge representations, which enables the model to learn richer sentence representations and perform much better.

(4) Knowledge embedding methods, including TransE, TransD, TransH, HolE, ComplEx, perform better than network embedding methods, including Deepwalk and LINE. As introduced in Section~\ref{section2}, NE methods basically learn the structural representations of KG, regardless of features of entities and relations, while KE methods embed the information from both the structure and the knowledge. Thus, KE methods are more effective for capturing the knowledge representations in knowledge graph application. 

(5)  Different kinds of KE methods achieve similar experimental results, but generally, translational distance models outperform semantic matching models in this task. TransE and TransH achieve the best performance in both frameworks and two datasets.

(6) Despite the small difference among these knowledge embedding methods, CKANN substantially improves the performance by leveraging GCN to adaptively learn contextualized knowledge representations, since pre-trained knowledge embeddings sometimes are too general and may not be sufficiently specific for certain problems. 

Overall, this experiment provides the answer to \textbf{RQ5}: \textit{Sophisticated knowledge representation learning methods actually can enhance the contribution of knowledge-based learning module in the proposed methods. In particular, using GCN to learn the adaptive knowledge representations is more effective than simply adopting those fixed pre-trained knowledge embeddings.}

\begin{table}
	\centering
	\caption{\label{table:knowledge-modlue} Comparisons on different knowledge modules}
	\begin{tabular}{lccccc}
		\toprule
		\multirow{2}{*}{Model}&\multicolumn{2}{c}{TREC QA(original)}&\multicolumn{2}{c}{Wiki QA} \\ 
		\cmidrule(lr){2-3}\cmidrule(lr){4-5}
		&MAP&MRR&MAP&MRR\\
		\midrule
		CKANN & \textbf{0.819} & \textbf{0.863} & \textbf{0.732} & \textbf{0.755} \\
		w/o KGs & 0.782 & 0.825  & 0.706 & 0.723\\
		\midrule
		- EviNets~\cite{DBLP:conf/acl/SavenkovA17}& 0.789 & 0.834  & 0.712 & 0.726\\
		- HR-BiLSTM~\cite{DBLP:conf/acl/YuYHSXZ17}& 0.792 & 0.841 & 0.710 & 0.725\\
		- QCN~\cite{DBLP:journals/ir/SawantGCR19}& 0.789 & 0.837 & 0.717 & 0.730\\
		\bottomrule 
	\end{tabular}
\end{table}

\subsubsection{Comparisons with Other Methods that Exploit External Knowledge} 
The proposed method jointly leverages the context information and external knowledge from knowledge graph. In order to evaluate the superiority of the proposed framework on exploiting knowledge information, we compare the proposed method with several methods which also leverage external knowledge by substituting the knowledge module with the following methods to learn the knowledge-based representations. Since these methods are not directly proposed for answer selection task, we incorporate them by implementing the core idea of the knowledge representation learning from these methods:
\begin{itemize}
    \item \textbf{EviNets}. \citet{DBLP:conf/acl/SavenkovA17} represent each piece of knowledge with a dense embeddings vector for measuring their relevance to the context via a simple dot product of context embeddings and knowledge embeddings. For a fair comparison, we employ pre-trained TransE embeddings for the knowledge as our methods, instead of learning from scratch, and omit the entity type information, which is absent in our problem setting.
    \item \textbf{HR-BiLSTM}. \citet{DBLP:conf/acl/YuYHSXZ17} combine word-level and knowledge-level relation embeddings with BiLSTM encoders for relation representational learning. In our implementation, we employ this method for our knowledge representational learning.
    \item \textbf{QCN}. \citet{DBLP:journals/ir/SawantGCR19} propose a query corpus network (QCN), which is a Siamese convolutional network, for measuring the relevance score between the context and each candidate entity. In our implementation, we use the score to conduct a weighted mean on the candidate entity embeddings for knowledge representational learning. 
\end{itemize}

The results are shown in the second part of Table~\ref{table:knowledge-modlue}. It can be observed that these methods can also improve the performance by exploiting the external knowledge information. However, the performance improvements (compared to ``w/o KGs") are limited, since these methods only conduct a shallow interaction learning between the context-based representations and the knowledge-based representations. This result demonstrates the superiority of the proposed contextualized knowledge representation learning module over existing alternative methods that can also integrate the context information with external knowledge in question answering.

\begin{figure*}
\centering
\includegraphics[width=\textwidth]{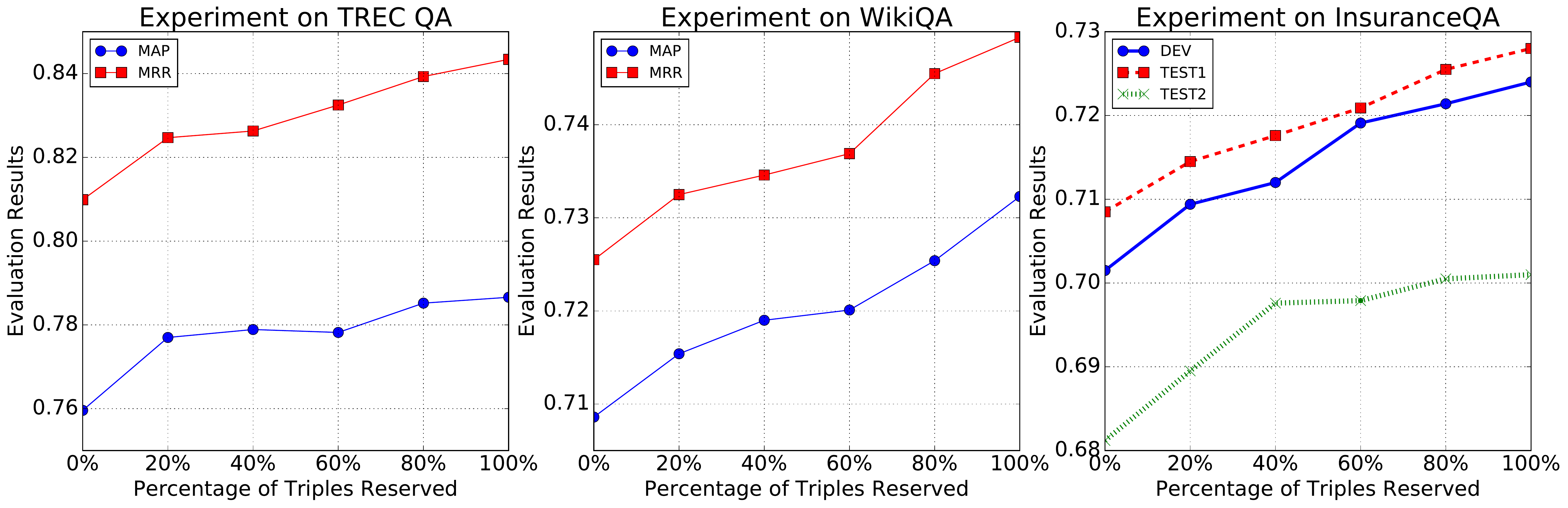}
\caption{Effect of KG completeness}
\label{completeness}
\end{figure*}

\subsection{Completeness of Knowledge Graph}\label{subsection55}
To analyze the performance of our model with respect to the completeness of the knowledge graph, we report the evaluation metrics on three datasets with incomplete subgraphs that  have only 20\%-80\% triples reserved.  Figure~\ref{completeness} shows the same tendency on three different datasets, which indicates that our model is robust and achieves excellent performance on the KG with different completeness. This result answers \textbf{RQ6}: \textit{As one may expect, training with more complete KG actually improves the overall performance, which also indicates the importance of background knowledge in QA.}

\subsection{Case Study}\label{subsection57}

Knowledge-aware attention mechanism provides an intuitive way to inspect the soft-alignment between the question and the answers by visualizing the attention weight from Equations~\ref{eq1} \&~\ref{eq2} for knowledge-aware self-attention and co-attention, respectively. We randomly choose one question-answer pairs from TREC QA dataset and visualize the attention scores predicted by attentive pooling scheme~\cite{DBLP:journals/corr/SantosTXZ16} and two proposed knowledge-aware attention mechanism. The color intensity indicates the importance degree of the words, the darker the more important. 

\begin{figure}
\includegraphics[width=0.7\textwidth]{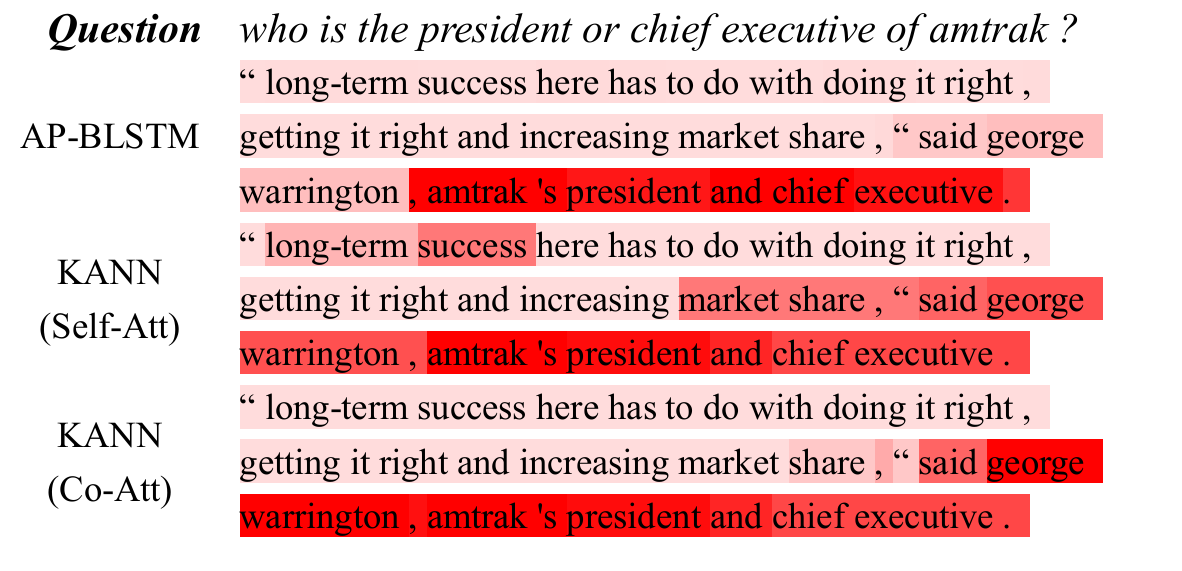}
\caption{An example of the visualization of knowledge-aware attention}
\label{att1}
\end{figure}

In Figure~\ref{att1}, we observe that attentive pooling scheme pays much attention to those words that are contextually related to the question, such as "amtrak", "president", "chief executive", while neglecting the knowledge beyond the context of the question like "george warrington". This limitation can be alleviated decently by knowledge-aware attention mechanism, since there is a strong correlation between "amtrak" and "george warrington" in the external knowledge graph. Thus, in knowledge-aware attention mechanism, "george warrington" is assigned with a higher weight. Compared with knowledge-aware co-attention, knowledge-aware self-attention also focuses on individual knowledge information in sentences, such as "success" and "market share", which leads to its worse performance than co-attention based methods.

\begin{figure}
	\centering
	\includegraphics[width=0.7\textwidth]{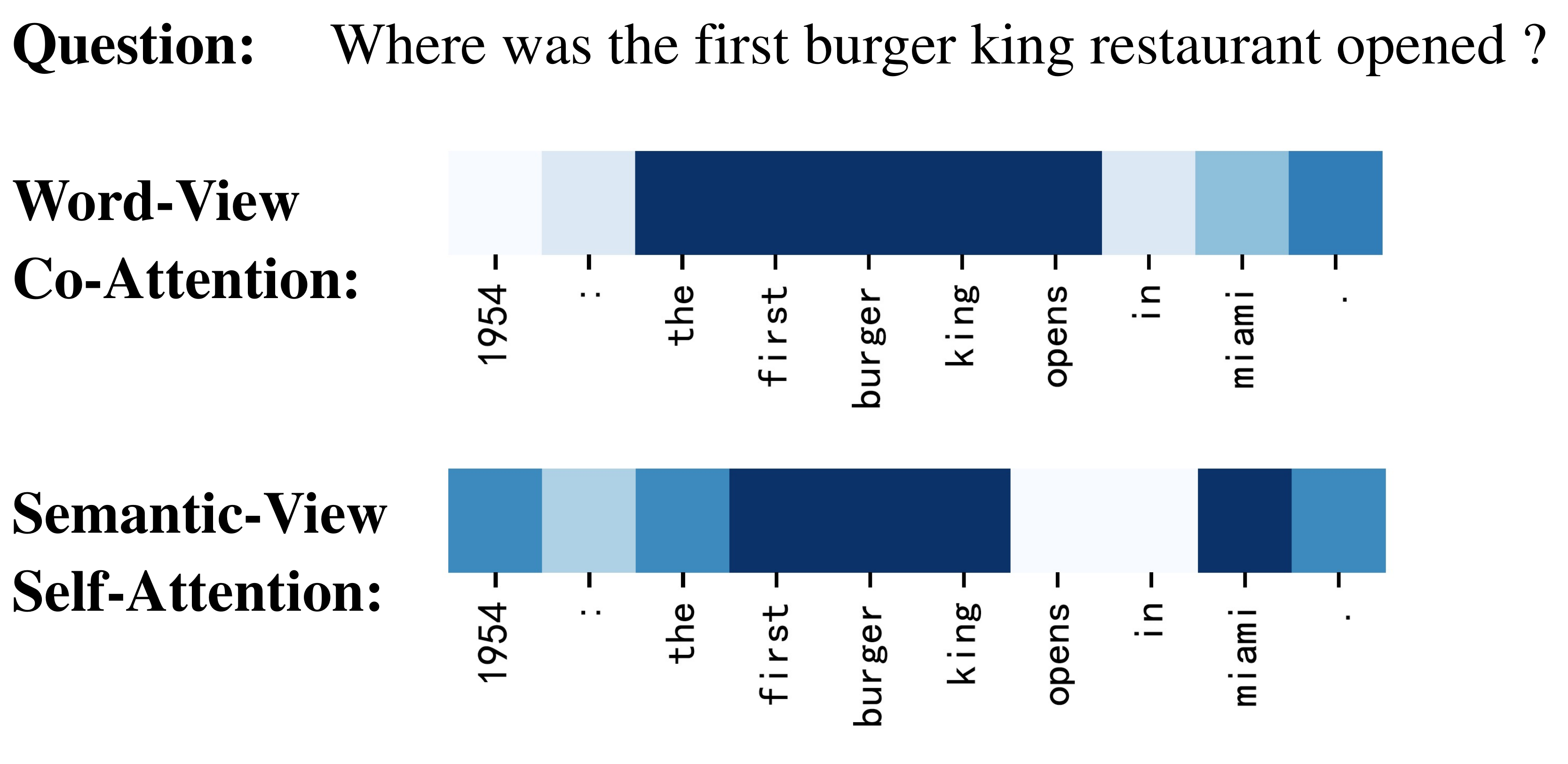}
	\caption{An example of the visualization of multi-view attention\label{fighre:case}}
\end{figure}

To better validate the performance of multi-view attention, we also randomly select a QA pair from the TREC QA dataset and visualize the word-view co-attention and semantic-view self-attention scores predicted by CKANN in Figure~\ref{fighre:case}. The color intensity indicates the importance degree of the words, the darker the more important.

We observe that word-view co-attention mechanism pays much attention to those words that are textually related to the question, such as ``the first'' and ``burger king'', but neglecting the knowledge relationship like ``miami''. This limitation can be alleviated decently by semantic-view self-attention mechanism, since there is a strong correlation between ``miami'' and ``burger king'' in the external KG. This result shows that multi-view attention scheme takes into account the important interaction information from different perspectives, which effectively contributes to the final performance.

\subsection{Error Analysis}
To better understand the failure modes of the proposed methods, we analyze 100 failure cases from four datasets (25 of each), and observe that the error cases can be categorized into the following groups for further studies.

\textbf{Information Imbalance:} ($\approx$ 40\%). Some QA pairs suffer a great imbalance on the provided information from the question and answer, resulting in difficulties in matching the question and answer. Such issues are more common in the error cases in YahooQA than other three datasets, since both the questions and answers in communitiy-based question answering are provided by non-expert users. Among them, about 60\% of them give a short and simple answer, such as ``Yes.", ``Sure." in YahooQA. While the rest only provide little information in the question, for which even human cannot determine the correct answer. One possible way to address this kind of failure is to introduce specific additional information, such as the question body in CQA~\cite{DBLP:conf/acl/WuWS18} or product reviews in E-Commerce QA~\cite{DBLP:conf/www/ZhangLDM20}, to balance the information from both the question and the answer.

\textbf{Mislabeling or Misspelling:} ($\approx$ 35\%). We attribute these failures to data issues. For instance, there are two candidate answers for the question “What years did Sacajawea accompany Lewis and Clark on their expedition?”: (i) “The coin honors the young woman and teenage mother who accompanied explorers Meriwether Lewis and William Clark to the pacific ocean in 1805”, and (ii) “In 1804, Toussaint was hired by Lewis and Clark, not for his own skills but for those of Sacagawea.” Despite the correctness of both answers, the first candidate answer is labelled as the correct answer, while the second is not. Besides, some ground-truth answers are misspelled. These issues are prevalent across four datasets.

\textbf{Knowledge Module Failure:} ($\approx$ 15\%). Compared to those context-based model, the issues are related to knowledge have been alleviated to a great extent. However, there are still some cases mismatched due to the failure of the knowledge module, including the absence of corresponding entities in the given knowledge graph, the mismatching of entity linking, and the failure of capturing correct relation between entities. Such issues are more common in two factoid QA datasets, TREC QA and WikiQA, as there is a larger demand of external knowledge in factoid QA. This indicates the necessity of knowledge graph related studies, including knowledge graph construction, entity linking, knowledge embedding, etc.

\textbf{Other Issues:} ($\approx$ 10\%). There are still some failure cases that we cannot clearly clarify the mismatch with certain reasons, such as the noise issue, the training issue, etc.

\section{Conclusions}\label{section6}
In this paper, we propose a knowledge-aware neural network (KNN) framework for answer selection, which effectively leverages external knowledge from KGs to enrich QA sentence representational learning. To summarize attention matrices from different sources, we propose two kinds of knowledge-aware attention mechanisms to tightly summarize the attention values between questions and answers. In order to dynamically and adaptively model the knowledge information, we further develop a Contextualized Knowledge-aware Attentive Neural Network (CKANN) for answer selection, which incorporates structure information from KGs via GCN and utilizes a multi-view attention mechanism to comprehensively attend the important information among context and knowledge. 

Experimental results show that the proposed framework can achieve substantial improvements over various state-of-the-art methods on several benchmark datasets. Besides, the ablation studies verify the effectiveness of some novel components in our framework, and the analysis also shows the broad practicability and universality of the proposed framework.

In the future, we are interested in incorporating entity type and description information to enhance knowledge representation, or introducing the relations in KGs to enrich the semantic relationship among nodes in the Entity Graph.


\bibliographystyle{ACM-Reference-Format}

\bibliography{sample-base}

\end{document}